\documentclass[lettersize,journal]{IEEEtran}
\usepackage{amsmath,amsfonts}
\usepackage{algorithmic}
\usepackage{algorithm}
\usepackage{array}
\usepackage[caption=false,font=normalsize,labelfont=sf,textfont=sf]{subfig}
\usepackage{textcomp}
\usepackage{stfloats}
\usepackage{url}
\usepackage{verbatim}
\usepackage{graphicx}
\usepackage{cite}
\usepackage{mathtools}
\usepackage{amssymb}
\usepackage{amsmath}
\usepackage{amsthm}
\usepackage{color}
\usepackage{graphicx}
\usepackage{booktabs}
\usepackage{comment}
\usepackage{caption}
\usepackage{multirow}
\usepackage{algorithm}
\usepackage{algorithmic}
\usepackage{newfloat}
\usepackage{listings}
\theoremstyle{plain}
\newtheorem{theorem}{Theorem}[section]

\theoremstyle{definition}
\newtheorem{definition}[theorem]{Definition}
\newtheorem{assumption}[theorem]{Assumption}
\theoremstyle{remark}
\newtheorem{remark}[theorem]{Remark}

\usepackage{newfloat}
\usepackage{listings}

\begin{document}

\title{Fast Propagation is Better: Accelerating Single-Step Adversarial Training via Sampling Subnetworks}

\author{Xiaojun Jia,
        Jianshu Li,
        Jindong Gu,
        Yang Bai
        and~Xiaochun Cao~\IEEEmembership{Senior Member,~IEEE}
\thanks{Manuscript received January 09, 2023; returned July 18, 2023;
revised August 26, 2023; accepted October 09, 2023. This work was Supported in part by the National Key R\&D Program of China (Grant No. 2022ZD0118100), in part by National Natural Science Foundation of China (No.62025604), in part by Beijing Natural Science Foundation (No.L212004), in part by Shenzhen Science and Technology Program (No. 20220016), in part by Nanyang Technological University (NTU)-DESAY SV Research Program under Grant 2018-0980. The associate editor
coordinating the review of this manuscript and approving it for publication
was Prof. Zhang, Haijun. (Corresponding author: Jindong Gu and Xiaochun Cao.) }
\thanks{ Xiaojun Jia is with
Cyber Security Research Centre @ NTU, Nanyang Technological University, Singapore, and also with Institute of Information Engineering, Chinese Academy of Sciences, Beijing 100093, China.
(e-mail: jiaxiaojunqaq@gmail.com).}
\thanks{Jianshu Li is with Ant Group, Beijing, China. (e-mail: jianshu.l@antgroup.com).} 

\thanks{Jindong Gu is with Torr Vision Group, University of Oxford. (e-mail: jindong.gu@outlook.com).} 


\thanks{Yang Bai is with Chengdu University of Information Technology, China.(e-mail: alicepub@163.com).}

\thanks{Xiaochun Cao is with School of Cyber Science and Technology, Shenzhen Campus, Sun Yat-sen University, Shenzhen 518107, China (e-mail: caoxiaochun@mail.sysu.edu.cn)}

}

\markboth{Manuscript for IEEE Transactions on Information Forensics and Security}%
{Shell \MakeLowercase{\textit{et al.}}: A Sample Article Using IEEEtran.cls for IEEE Journals}


\maketitle
\begin{abstract}
Adversarial training has shown promise in building robust models against adversarial examples. A major drawback of adversarial training is the computational overhead introduced by the generation of adversarial examples. To overcome this limitation, adversarial training based on single-step attacks has been explored. Previous work improves the single-step adversarial training from different perspectives, \textit{e.g.,} sample initialization, loss regularization, and training strategy. Almost all of them treat the underlying model as a black box. In this work, we propose to exploit the interior building blocks of the model to improve efficiency. Specifically, we propose to dynamically sample lightweight subnetworks as a surrogate model during training. By doing this, both the forward and backward passes can be accelerated for efficient adversarial training. Besides, we provide theoretical analysis to show the model robustness can be improved by the single-step adversarial training with sampled subnetworks. Furthermore, we propose a novel sampling strategy where the sampling varies from layer to layer and from iteration to iteration. Compared with previous methods, our method not only reduces the training cost but also achieves better model robustness. Evaluations on a series of popular datasets demonstrate the effectiveness of the proposed FB-Better. Our code has been released at https://github.com/jiaxiaojunQAQ/FP-Better.

\end{abstract}

\section{Introduction}
Deep neural networks(DNNs) have been known to be vulnerable to adversarial examples (AEs)~\cite{szegedy2013intriguing,goodfellow2014explaining,li2019nattack,dong2018boosting,wang2021enhancing,bai2021targeted,fan2020sparse,liu2022watermark,liang2022large}, which are generated via adding imperceptible perturbations to benign data. The vulnerability of DNNs to adversarial examples poses potential threats to DNN-based real-world applications. To address the risk brought by adversarial examples, many attack and defense methods have been proposed. After the attack-defense arms race in the past years~\cite{athalye2018obfuscated}, adversarial training (AT) ~\cite{goodfellow2014explaining,madry2017towards,wong2020fast,DBLP:journals/tifs/HuangJGLP22,li2022semi,jia2022adversarial,DBLP:conf/nips/MaoCDZQYLZ022,mao2022towards} becomes one of the most effective methods to enhance adversarial  robustness against adversarial examples.
\begin{figure}[t]

        \centering
        \includegraphics[width=0.95\linewidth]{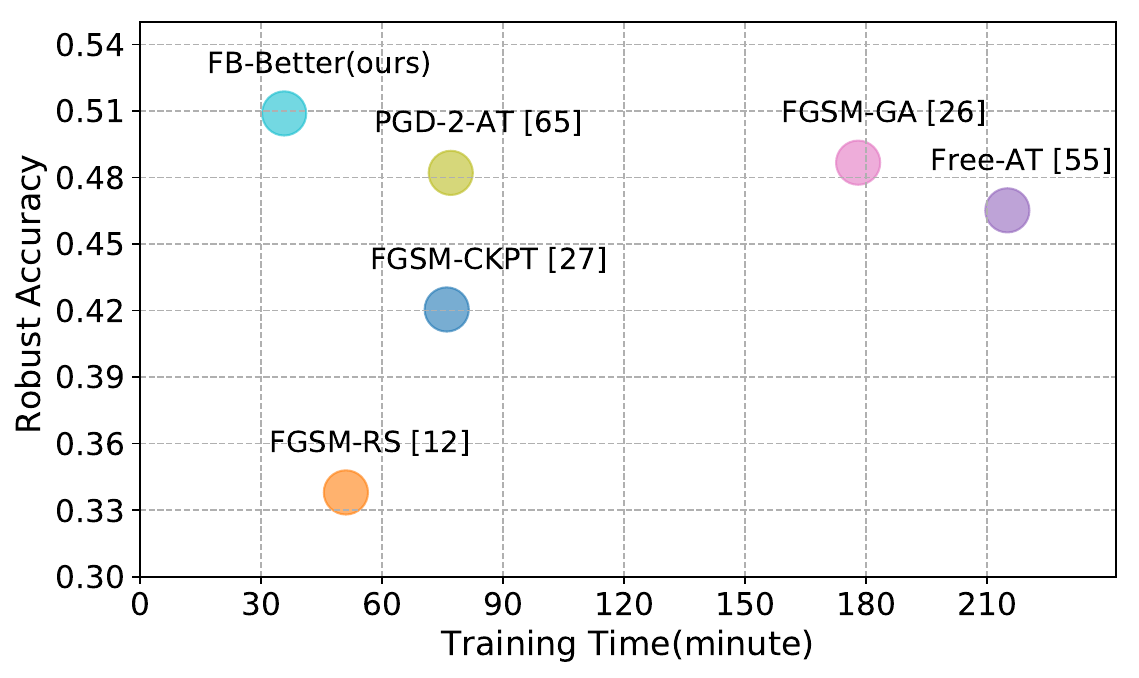}
        \label{fig:time_acc}

  \caption{ Training time and robust accuracy under PGD-10 of series of single-step adversarial training methods using ResNet18 with the best checkpoint on the CIFAR-10 image dataset under $\ell_{\infty}= 8/225$. $x$-axis illustrates the training time. And $y$-axis illustrates the robust accuracy under PGD-10. 
  }
\label{fig:time_acc}
\end{figure}
Adversarial training boosts the adversarial robustness by injecting adversarial examples into training data. The injected adversarial examples are created online during the training process, which is computationally expensive. For example, the training process can be N (\textit{e.g.,} 3-40) times longer than the standard training process when the popular multi-step attack PGD is applied to create adversarial examples~\cite{madry2017towards,zhang2019you,DBLP:conf/mm/00010JMMLP21,dai2021parameterizing,bai2021improving,cui2021learnable,DBLP:conf/nips/ZieglerNCBSLSNW22,duan2021advdrop,duan2021adversarial}.

\begin{figure*}[t]
\begin{center}
 \includegraphics[width=1\linewidth]{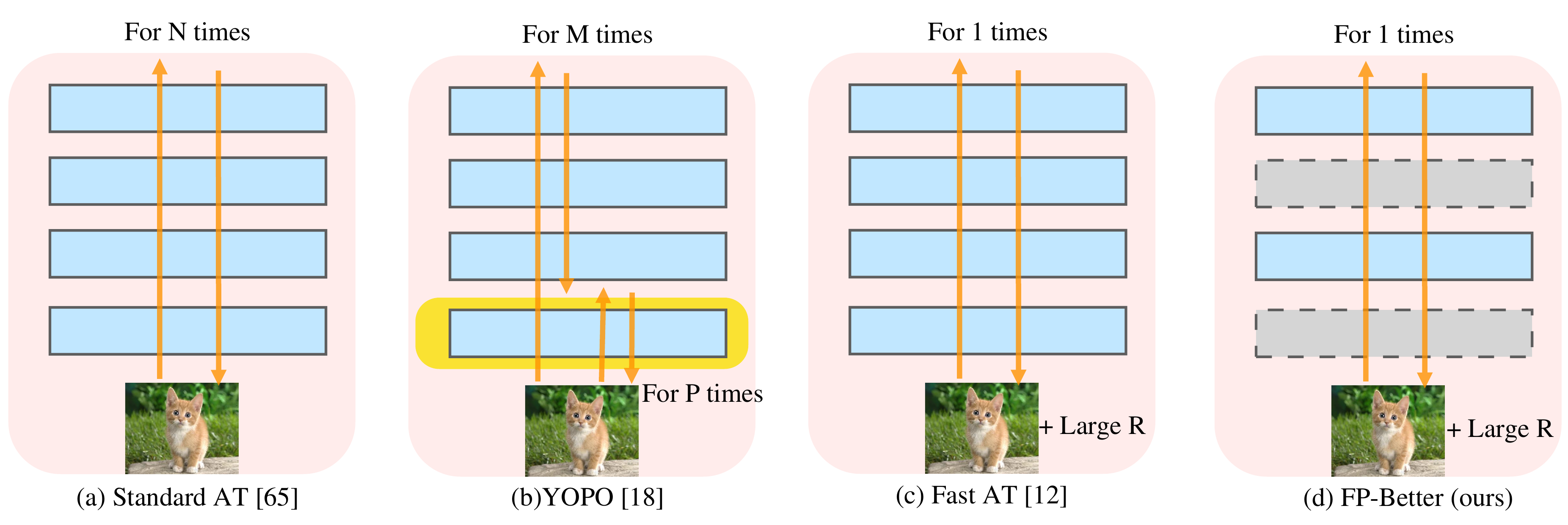}
\end{center}
        \caption{Overview of our FP-Better. The standard adversarial training requires N times forward and backward passes for each mini-batch. To reduce the computational cost, the work YOPO constrains part of passes only in the first layer of the model. A recent work about fast adversarial training shows adversarial training with the single-step attack can also achieve competitive robustness when large noises are added to inputs as a random initialization. Our FP-Better makes each forward and backward pass more efficient by sampling a lightweight subnetwork in each training iteration.}
        \label{fig:fp_better}
\end{figure*}

To address the computational overhead introduced by the generation of adversarial examples in adversarial training, single-step adversarial training is proposed where the single-step attack is adopted to create adversarial examples~\cite{wong2020fast,andriushchenko2020understanding,kim2020understanding,sriramanan2020guided,sriramanan2021towards,xiong2022stochastic,yuan2022adaptive,zhu2022toward,li2022subspace,jia2022boosting,vivek2020single}. Concretely, the popular adversarial training generates adversarial examples by using the fast gradient sign method (FGSM)~\cite{szegedy2013intriguing}, dubbed FGSM-AT. Though FGSM-AT improves the training efficiency and model robustness against adversarial examples, it can encounter catastrophic overfitting during training, \textit{i.e.,} the model robustness accuracy against multi-step attacks suddenly drops to 0\% after a few training epochs. In recent years, many approaches have been proposed to improve the attack effectiveness of adversarial examples and mitigate the catastrophic overfitting in single-step adversarial training, such as designing training schedules~\cite{wong2020fast,kim2020understanding,sriramanan2021towards,li2022subspace,jia2022boosting},
regulariziong training processes~\cite{andriushchenko2020understanding,jia2022prior}. The model is treated as a black box in these approaches.

The work~\cite{zhang2019you} makes the first exploration to design DNN-specific algorithms to accelerate adversarial training. As illustrated in Fig.~\ref{fig:fp_better}, they propose to constrain part of forward and backward passes only in the first layer of the model, which is more efficient than standard multi-step adversarial training shown in Fig.~\ref{fig:fp_better}. They show You Only Propagation Once (\textbf{YOPO}), and the gradients of the next rounds can be obtained by propagating the previous gradients through the first layer. However, the attack effectiveness of the obtained adversarial examples is very limited since part of the gradients are obtained from the first layer. In this work, we propose to sample a subnetwork as a surrogate model to compute gradients. The forward and backward passes on the subnetworks are much more efficient than those on the original model. We show Fast Propagation is Better, dubbed \textbf{FP-Better}. 

An overview of our method is shown in Fig.~\ref{fig:fp_better}. Specifically, we sample a subnetwork from the underlying model as a surrogate model in each training iteration. In this work, we also provide a theoretical analysis to show adversarial training with the sampled subnetworks can improve the robustness of the underlying models. Based on the investigation, we further propose a novel sampling strategy. We show our FP-better with the proposed sampling strategy achieves state-of-the-art performance under attack evaluation. 

Recent work towards understanding single-step adversarial training reveals that catastrophic overfitting phenomenon can be well mitigated with appropriate regularization methods, \emph{e.g.,} with large perturbation range~\cite{wong2020fast}, with dropout strategies~\cite{vivek2020single} or with gradient alignment regularization~\cite{andriushchenko2020understanding}. Our method can also be seen as a regularization method where we sample a subnetwork for each training iteration. In detail, the proposed method samples sub-networks by randomly dropping the redundant repeated blocks, which can reduce the dependence between network layers, to prevent  catastrophic overfitting for fast adversarial training.
Hence, the proposed method not only prevents catastrophic overfitting but also improves adversarial training for fast adversarial training, which is also well supported by empirical experiments. As shown in Fig.~\ref{fig:time_acc} which represents the training time and robust accuracy of a series of single-step adversarial training methods, it can illustrate that the proposed FB-Better significantly improves the training efficiency and adversarial robustness. It is particularly noteworthy that our FB-Better is faster than {FGSM-RS~\cite{wong2020fast}}. It only requires 70\% training time of FGSM-RS~\cite{wong2020fast} which is the fastest adversarial training method of the previous.  

Our contributions can be summarized as follows:
\begin{itemize}
     \item We propose to accelerate single-step adversarial training from the perspective of opening the black-box model, namely, via sampling lightweight subnetworks as surrogate models.
     
    \item A theoretical analysis is provided to show the model robustness can be improved by the single-step adversarial training on the surrogate subnetworks sampled from the underlying model. 
    
    \item We propose a novel sampling strategy to sample subnetworks from the underlying model as surrogate models during training where the sampling varies from layer to layer and from iteration to iteration.

    \item Experiments and analyses on four standard datasets are conducted to demonstrate the effectiveness of the proposed method. The proposed single-step adversarial training method achieves state-of-the-art performance.
\end{itemize}

\section{Related Work}
 At first, we introduce the adversarial attack methods to generate adversarial examples for robustness evaluation and adversarial training defense methods which include multi-step and single-step adversarial training methods to defend against adversarial examples. We introduce the single-step adversarial training methods from three  perspectives, \emph{i.e.,} sample initialization, loss regularization, and training strategy. 
\subsection{Adversarial Attack Methods}
Szegedy  \emph{et al}. \cite{szegedy2013intriguing} are the first to discover the existence of adversarial examples. Goodfellow \emph{et al}. \cite{goodfellow2014explaining} proposed to adopt the fast gradient sign method (FGSM) which makes use of the model gradient for the generation of adversarial examples. To improve the performance of FGSM, Moosavi-Dezfooli \emph{et al}. \cite{moosavi2016deepfool} proposed a simple and accurate method to fool DNNs, called DeepFool. It exploited an iterative linearization of the classifier to generate adversarial examples. Then Tram{\`e}r \emph{et al}. \cite{tramer2017ensemble} proposed to add a randomization step to FGSM to generate adversarial examples, called R+FSGM.
Later, Madry \emph{et al}. \cite{madry2017towards} proposed to adopt  projected gradient descent (PGD) which adopts the model gradient iteratively to generate adversarial examples. Carlini \emph{et al}. \cite{carlini2017towards} proposed several optimization-based attack methods to attack DNNs, which are widely used to evaluate the model robustness, called C\&W.  A series of adversarial attack methods \cite{dong2019evading,lin2019nesterov,xie2019improving} adopt various input transformations to improve the attack transferability of adversarial examples. Some adversarial attack methods~\cite{brendel2017decision,ilyas2018black,li2019nattack,chen2020boosting} are conducted to generate adversarial examples in the black-box setting, \textit{i.e.,} attackers have no access to DNNs. Croce  \emph{et al}. \cite{croce2020reliable} explored the limitations of PGD and proposed two improved adversarial attack methods (APGD-DLR, APGD-CE) based on PGD. And then combining with other two adversarial attack methods (FAB~\cite{croce2020minimally} and Square~\cite{andriushchenko2020square}), they proposed a parameter-free ensemble of attacks to evaluate the model robustness, called AutoAttack (AA). In this paper, we make use of the widely used attack methods, which include PGD, C\&W, and AA to evaluate the adversarial robustness of the proposed method.

\subsection{Adversarial Training Methods}
Adversarial training (AT) methods \cite{zhang2019theoretically,wang2019improving,roth2019adversarial,croce2020minimally,DBLP:conf/icml/Yu0SY0GL22} have been proved to be an  effective defense method to defend against adversarial examples. 
Madry \emph{et al}. \cite{madry2017towards} formulate the adversarial training as a problem of minimax optimization. It is formulated as: 
\begin{equation}
\min _{\boldsymbol{\theta}} \mathbb{E}_{(\mathbf{x}, {y}) \sim \mathcal{U}}[\max _{\boldsymbol{\delta} \in \Omega} \mathcal{L}(f(\mathbf{x}+\boldsymbol{\delta}, \boldsymbol{\theta}), {y})],
\end{equation}
where  $f(\cdot)$ is the underlying mode, $\mathcal{U}$ is a data distribution. $\mathbf{x}$ is the clean image, $y$ is the corresponding ground truth label, $ \mathcal{L}$ is the loss of a deep network with the parameter $\boldsymbol{\theta}$, $\boldsymbol{\delta}$ is the
adversarial perturbation generated by adversarial attack methods, and $\Omega=\{\boldsymbol{\delta}:\|\boldsymbol{\delta}\| \leq \epsilon\}$ is a threat bound with  the maximum perturbation strength $\epsilon$. The adversarial training methods can be roughly divided into multi-step adversarial training methods  and single-step adversarial training methods depending on how the adversarial perturbation $\boldsymbol{\delta}$ is generated. Multi-step adversarial training makes use of  multi-step attacks for the adversarial perturbation generation to conduct adversarial training. One classic adversarial attack method is PGD. It is formulated as:
\begin{equation}
\boldsymbol{\delta}^{t+1}_{adv}=\Pi_{[-\epsilon, \epsilon]}\left[\boldsymbol{\delta}^{t}_{adv}+\alpha \cdot \operatorname{sign}\left(\nabla_{\mathbf{x}} \mathcal{L}\left(f(\mathbf{x}+\boldsymbol{\delta}^{t}_{adv},\boldsymbol{\theta} ), y  \right)\right)\right],
\end{equation}
where $\Pi_{[-\epsilon, \epsilon]}$ represents a projection operation which projects the input to the range of $[-\epsilon, \epsilon]$, $\boldsymbol{\delta}^{t}_{adv}$ is the generated adversarial perturbation during the $t$-th iteration, and $\alpha$ is the step size. A series of advanced multi-step adversarial training methods~\cite{wang2019improving,lee2020adversarial,bai2021improving,wang2021convergence} are proposed from different perspectives based on PGD to improve model robustness. Although these methods have achieved excellent performance in improving model robustness, they require a lot of computational time to generate adversarial examples for adversarial training. 
\par To improve the training efficiency of multi-step adversarial training, a series of adversarial training variants~\cite{wong2020fast,kim2020understanding,andriushchenko2020understanding} have been proposed from different perspectives, \emph{i.e.,} sample initialization, loss regularization, and training strategy. 

\subsubsection{Sample initialization}
Shafahi  \emph{et al}. \cite{shafahi2019adversarial} propose to adopt the single model gradients to simultaneously update the adversarial perturbation and  the model weights and then conduct multi-step adversarial training, called Free-AT. In this way, it speeds up AT by reducing redundant calculations during backpropagation. To further improve the training efficiency of adversarial training, Wong \emph{et al}. \cite{wong2020fast} recommend combining a random initialization with FGSM to generate adversarial examples for single-step adversarial training. Then they adopt early stopping to prevent catastrophic overfitting and achieve comparable model robustness to the primary PGD-AT~\cite{madry2017towards}, called FGSM-RS.
The adversarial perturbation of FGSM-RS is formulated as:
\begin{equation}
\boldsymbol{\delta}_{adv}=\Pi_{[-\epsilon, \epsilon]}\left[\boldsymbol{\varphi}+\alpha \cdot \operatorname{sign}\left(\nabla_{\mathbf{x}} \mathcal{L}\left(f(\mathbf{x}+\boldsymbol{\varphi}, \boldsymbol{\theta}), y  \right)\right)\right],
\label{eq:FGSM_RS}
\end{equation}
where $\boldsymbol{\varphi} \in \mathbf{U}(-\epsilon, \epsilon)$ is a random initialization, $\mathbf{U}$ represents a uniform distribution. Note that FGSM-RS is the fastest method for adversarial training.  
The algorithm of FGSM-RS is summarized in Algorithm \ref{alg:FGSM_RS}. {Moreover, Jia  \emph{et al}.~\cite{jia2022boosting} propose to adopt an additional generative network to generate a learnable sample initialization for fast adversarial training to further improve adversarial robustness. Then Jia  \emph{et al}.~\cite{jia2022prior} propose a priori-guided sample initialization, which requires additional storage memory, to boost adversarial robustness. Although these methods can effectively improve adversarial robustness, they require more training burden.}

\subsubsection{Loss regularization}
Andriushchenko \emph{et al}. \cite{andriushchenko2020understanding} demonstrate that only using the random
initialization delays catastrophic overfitting and does not prevent it. Then they propose a gradient regularization method (GradAlign) for FGSM-RS to prevent catastrophic overfitting, called FGSM-GA. Sriramanan \emph{et al}.~\cite{sriramanan2020guided} adopt a guided regularization method of function smoothing to boost adversarial robustness. Moreover, Sriramanan \emph{et al}.~\cite{sriramanan2021towards} propose to make use of a Nuclear-Norm regularization method for function smoothing to further improve adversarial robustness. Although these loss regularization methods can improve the robustness of fast adversarial training, they are to equip massive training time to calculate the proposed loss regularization. In this paper, we focus on how to improve adversarial robustness without extra training time. Hence, we employ typical fast adversarial training of loss regularization, \emph{i.e.,} FGSM-GA to conduct comparative experiments. {Although applying regularization in fast adversarial training can significantly prevent catastrophic overfitting and improve robustness, extra training time is required to compute the regularization.}

\subsubsection{Training strategy}
Kim \emph{et al}. \cite{kim2020understanding} claim that  catastrophic overfitting is caused by FGSM-RS only using adversarial examples with the maximum perturbation instead of ones in the adversarial direction. Then they propose a simple and effective training strategy method to select the optimal step size to generate adversarial examples for training. Although they analyze the reason for catastrophic overfitting from different perspectives and propose their own methods to prevent it, they require more computing time than FGSM-RS. Dhillon \emph{et al}.~\cite{dhillon2018stochastic} propose a mixed training strategy, \emph{i.e.,} Stochastic Activation Pruning, to defend against adversarial examples, called SAP. It randomly prunes a random activation to achieve adversarial robustness. Li \emph{et al}.~\cite{li2020towards} propose a ticket training strategy to obtain a robust model by pruning a non-robust model based on the lottery ticket hypothesis~\cite{frankle2018lottery}, called Ticket. Vivek \emph{et al}.~\cite{vivek2020single} claim that the original fast adversarial training achieves the pseudo robustness by the gradient masking effect and propose a dropout training strategy for fast adversarial training to obtain the real adversarial robustness, called Dropout. These training strategy-based methods effectively prevent catastrophic overfitting. 
However, the brought robustness improvement is only limited.

\par  Although the above fast adversarial training methods effectively enhance adversarial robustness from different perspectives, most of them treat the underlying model as a black box.
 Zhang \emph{et al.} ~\cite{zhang2019you} redefine adversarial training as a discrete-time differential game and then indicate that the generation of adversarial examples is only coupled to the weights of the first layer by analyzing Pontryagin's Maximum Principle. They make the first exploration to design DNN-specific algorithms to accelerate adversarial training. In detail, they constrain part of forward and backward passes only in the first layer of the model, called You Only Propagation Once (YOPO). YOPO avoids multiple calculations for full forward and backward propagation, which is more efficient than standard multi-step adversarial
training. But the attack effectiveness of the adversarial examples is limited since the gradients only are obtained from the first layer, which restricts further adversarial robustness improvement. In this paper, we exploit the interior building blocks of the model to improve efficiency and propose a novel sampling training strategy to boost the adversarial robustness of fast adversarial training.

\begin{algorithm}[t]
\caption{FGSM-RS}
\label{alg:FGSM_RS}
\begin{algorithmic}[1] 
\REQUIRE
The whole epoch $M$, the attack step size $\alpha$, the adversarial perturbation $\epsilon$, the label $y$, the clean image $\mathbf{x}$, the database size $N$ and the parameters of the trained network $\boldsymbol{\theta}$.

\FOR{$j=1,...,M$}
\FOR{$i=1,...,N$}
\STATE $\boldsymbol{\varphi} =\mathbf{U}(-\epsilon ,\epsilon) $
\STATE $ \boldsymbol{\delta}_{adv}=\Pi_{[-\epsilon, \epsilon]}\left[\boldsymbol{\varphi}+\alpha \cdot \operatorname{sign}\left(\nabla_{\mathbf{x}} \mathcal{L}\left(\mathbf{x}_{i}+\boldsymbol{\varphi}, y_{i} ;\boldsymbol{\theta}  \right)\right)\right]$
\STATE $ \boldsymbol{\theta} \leftarrow \boldsymbol{\theta} -\nabla_{\boldsymbol{\theta}}\mathcal{L}(\mathbf{x}_{i} + \boldsymbol{\delta}_{adv},y_{i};\boldsymbol{\theta})  $
\ENDFOR
\ENDFOR
\end{algorithmic}
\end{algorithm}

\section{The Proposed Approach}
In this section, we first introduce the framework of the proposed method in Sec.~\ref{Framework}. Then we provide the theoretical analysis to verify the effectiveness of the proposed method in Sec.~\ref{Generalization_analysis}. Moreover, we propose a novel sampling strategy to sample subnetworks for single-step adversarial training in Sec.~\ref{Strategy}.


\subsection{Framework of Our FB-Better for Single-step Adversarial Training }
\label{Framework}
In each training iteration, we sample a subnetwork $f'(\cdot)$ from the underlying model $f(\cdot)$ to generate adversarial examples and train the sampled subnetwork on the generated adversarial examples. The sampled subnetwork $f'(\cdot)$ varies from iteration to iteration. The adversarial training on $f(\cdot)$ can be formulated as: 
\begin{equation}
\min _{\boldsymbol{\theta}} \mathbb{E}_{(\mathbf{x}, {y}) \sim \mathcal{U}}[\max _{\boldsymbol{\delta}_{adv} \in \Omega} \mathcal{L}(f(\mathbf{x}+\boldsymbol{\delta}_{adv}, \boldsymbol{\theta}), {y})].
\end{equation}
In $i$-th training iteration, we apply FGSM to the mini-batch $\mathbf{x}_i$ on the sampled subnetwork $f'_i(\cdot)$ to generate adversarial examples to conduct adversarial training for the subnetwork. It can be defined as:
\begin{equation}
\min _{\boldsymbol{\theta}} \max _{\boldsymbol{\delta}_{adv} \in \Omega} \mathcal{L}(f'_i(\mathbf{x}_i+\boldsymbol{\delta}_{adv}, \boldsymbol{\theta}), {y}).
\end{equation}
The adversarial perturbation $\boldsymbol{\delta}_{adv}$ is the core to improve the adversarial robustness. In this work, it is generated by single-step attack methods (FGSM) on the subnetwork.
It can be defined as:
\begin{equation}
\boldsymbol{\delta}_{adv}=\Pi_{[-\epsilon, \epsilon]}\left[\boldsymbol{\varphi}+\alpha \cdot \operatorname{sign}\left(\nabla_{\mathbf{x}} \mathcal{L}(f'_i(\mathbf{x}_i+\boldsymbol{\varphi}, \boldsymbol{\theta}), y) \right)\right].
\label{eq:FGSM_SD}
\end{equation}
Then the generated adversarial examples are used to train the subnetwork $f'_i(\cdot)$. During training, the model parts are trained once selected. The whole model $f(\cdot)$ is used as the final robust model. 
The effectiveness of the proposed approach is verified by both theoretical and empirical analysis.

\subsection{Theoretical Analysis}
\label{Generalization_analysis}
For our algorithm $\mathcal A$ learns a hypothesis $h$ on the training sample set $S$, the expected risk $\mathcal R(h)$ and empirical risk $\hat{\mathcal R}(h)$ are defined as follows,
\begin{gather*}
	 \mathcal R(h) = \mathbb E_{Z} l(h, Z),~~
	\hat{\mathcal R}_S(h) = \frac{1}{N} \sum_{i=1}^N l(h, z_i).
\end{gather*}
Then, we can obtain a generalization bound of our algorithm based on He \emph{et al}. \cite{he2020robustness}.

\begin{theorem}
\label{thm:high_probability_privacy}
Suppose one employs SGD for adversarial training. $L_{ERM}$ is the maximal gradient norm in ERM. Also, suppose the whole training procedure has $T$ iterations. Then, the algorithm $\mathcal A$ has a high-probability generalization bound as follows. Specifically, the following inequality holds with probability at least $1 - \gamma$:
\begin{align*}
\label{eq:high_probability_privacy}
    \mathbb{E}\mathcal R(h) - \mathbb{E}\hat{\mathcal R}_S(h) 
   \le  c (M(1-e^{-\varepsilon}+e^{-\varepsilon}\delta) \log{N}\log{\frac{N}{\gamma}} 
   \\ +\sqrt{\frac{\log 1 / \gamma}{ N}} )\\, 
\end{align*}
where
 \begin{equation}
 \label{eq:epsilon_iteration}
\varepsilon = \varepsilon_0\sqrt{2 T \log \frac{N}{\delta'}} +T \varepsilon_0 (e^{\varepsilon_0} - 1),~~
\delta =  \frac{\delta'}{N},~~
\end{equation}
 \begin{equation}
 	\label{eq:epsilon_whole}
		\varepsilon_0= \frac{2 L_{ERM}}{N b} \prod_{i = 1}^d \left( \frac{ \max_{\theta, x, y} \left \| \nabla_{\theta} \mathcal L_{\text{adv}}^i \right \|}{\max_{\theta, x, y} \left \| \nabla_{\text{grad}}^i \right \|} \right), 
\end{equation}

$\nabla_{\theta} \mathcal L_{\text{adv}}^i$ and $\nabla_{\text{grad}}^i$ are the $i$-th entry of $\nabla_{\theta} \mathcal L_{\text{adv}}$ and $\nabla_{\text{grad}}$, respectively, defined for the $i$-th layer, $d$ is the depth, $\delta'$ is a positive real, $\tau$ is the batch size, $I$ is the robustified intensity,
\begin{equation*}
I = \frac{ \max_{\theta, x, y} \left \| \nabla_{\theta} \max_{\Vert x^\prime - x \Vert \leq \rho} l (h_{\theta} (x^\prime), y) \right \|}{\max_{\theta, x, y} \left \| \nabla_{\theta} l (h_{\theta} (x), y) \right \|},
\end{equation*}
$b$ is the Laplace parameter, $\gamma$ is an arbitrary probability mass, $M$ is the bound for loss $l$, $N$ is the training sample size, $c$ is a universal constant for any sample distribution,  and the probability is defined over the sample set $S$.
\end{theorem}

The proof is given in the \textbf{appendix}.

\begin{remark}
Eq. (\ref{eq:epsilon_iteration}) characterizes the influence of every layer, where
$    \frac{ \max_{\theta, x, y} \left \| \nabla_{\theta} \mathcal L_{\text{adv}}^i \right \|}{\max_{\theta, x, y} \left \| \nabla_{\text{grad}}^i \right \|}
$
characterizes the influence from the $i$-th layer.
\end{remark}

When subnetworks are sampled during training, some layers of the underlying model are randomly dropped. Thus, the term $\varepsilon_0$ decreases; and therefore, the generalization bound of $\mathbb{E}\mathcal R(h) - \mathbb{E}\hat{\mathcal R}_S(h)$ decreases. This suggests that training with dynamically sampled subnetworks can improve the generalization, when the adversarial robustness is fixed. 

\subsection{A Novel Sampling Strategy}
\label{Strategy}
In the first two subsections, we describe adversarial training with dynamically sampled subnetworks. In this subsection, we propose a novel sampling strategy to sample subnetwork. The current SOTA model architectures consist of repeated blocks. The sampling can be implemented by dropping the selected blocks. In detail, during training, some layers are dropped during both the forward and backward passes. Specifically, during both the forward and backward passes, some residual blocks are skipped and the shortcut path is kept. During the forward pass of training, each layer has a probability of being dropped or skipped. During the backward pass, only the layers that are not dropped during the forward pass contribute to the gradient computation.
Actually, all the blocks can be dropped with a certain probability. For a repeated block where the input and output of the block are different, the residual part can be safely dropped since the feature dimensions are the same in both the input and output of the block. For a bottleneck block where the input and output of the block are different, the residual part can also be dropped in the same way. The size can still be held since the skip connection also includes a downsampling operation (\emph{i.e.,} a convolutional operation). The feature sizes of all the blocks are kept the same before and after the dropping operations. Hence, the linear classifier can be always kept without changing the dimensions.
Our sampling strategy is implemented by sampling blocks from both temporal and spatial dimensions, \emph{i.e.,} the sampling varies from layer to layer and from iteration to iteration. Specifically, the spatial dimension indicates that the sampling strategy in the spatial dimension is related to the model architecture, \emph{i.e.,} modules of different depths have different sampling probabilities. Higher modules have higher sampling strategies, which vary linearly. The temporal dimension indicates that the sampling strategy in the temporal dimension is related to the training time of the model, \emph{i.e.,} as the training continues, we gradually increase the sampling probabilities of each block. The closer to the later stage of training, the higher the sampling probability of each block.

\subsubsection{Sampling strategy in the spatial dimension} The proposed sampling strategy on the spatial dimension is simple. Intuitively, low-level features which are extracted by the earlier layers are used by later layers. They need to be more reliably presented, \textit{i.e.,} the earlier the layer is, the more it needs to be preserved Huang \emph{et al}. \cite{huang2016deep}. Hence, we propose a simple sampling strategy where the sampling probability decreases linearly with blocks.
The sampling probability of the top blocks is set to $p_{min}$. The sampling probabilities of all blocks can be defined as:
\begin{equation}
p_{\ell}=1-\frac{\ell}{L}\left(1-p_{min}\right),  \quad 1 \leq \ell \leq L
\end{equation}
where L is the number of blocks of the whole network.

\subsubsection{Sampling strategy in the temporal dimension}
\label{temporal_spatial} 
We now present the sampling strategy on the temporal dimension. As for the sampled subnetwork for AT, there is an efficiency-performance trade-off, \textit{i.e.,} the more training time it takes, the more robust the model is. We can understand this phenomenon in terms of the bias-variance trade-off principle\cite{yang2020rethinking,hayou2021regularization}. Specifically, we use the $L$ to represent the number of blocks  of the whole network and the $\widetilde{L}$ to represent the number of blocks of the subnetwork.
When $\widetilde{L} / L \rightarrow 0$, the model bias increases (and the model variance decreases). The increase of the bias inevitably deteriorates the adversarial robustness performance. In other words, reducing the training time can decrease the adversarial robustness. Moreover, the variance and bias of the model change dynamically as training progresses. During the whole training stage, only using a sampling strategy on the spatial dimension with the fixed sampling probability $p_{\ell}$ may limit the performance improvement.

 \par To further improve the adversarial robustness and training efficiency, we propose a dynamically changing sampling rate to conduct adversarial training. In detail, at the beginning of training, we adopt the subnetwork with the shallow depth (small sampling rate $p_{\ell}$ to conduct adversarial training). As the training continues, we gradually increase the depth of the model. In this way, the proposed method not only further improves the robustness of the model but also further reduces the computation time. The core of the proposed method is when to adjust the sampling rate. We design a simple yet effective adjusting criterion by the cumulative adversarial training loss over a certain period of time. It can be defined as:  
 \begin{equation}
\varpi=\sum_{i=1}^{N}\mathcal{L}_{\text{cur}}(f(\mathbf{x}_{i}^{adv}, \boldsymbol{\theta}_{i}^{\text{cur}}), y)-\sum_{i=1}^{N}\mathcal{L}_{\text{pre}}(f(\mathbf{x}_{i}^{adv}, \boldsymbol{\theta}_{i}^{\text{pre}}), y),
\end{equation}
where $N$ represents the iterative training times over a certain period of time, $\mathcal{L}_{\text{cur}}$ represents the cumulative adversarial training loss for the current certain period of time, and $\mathcal{L}_{\text{pre}}$ represents the cumulative adversarial training loss for the previous certain period of time. If $\varpi>0$, this means that the model with current depth can continue to be trained to improve robustness performance. Otherwise, the model with current depth may have achieved the upper limit of robustness performance. The depth of the current model needs to be increased, \textit{i.e.,} increasing the sampling rate $p_{\ell}$.
\par To keep the sampling strategy on the spatial dimension, we change the linear decaying sampling probability from $[1,p_{min}]$ to $[1,\tilde{p}_{min}]$ ($\tilde{p}_{min}>p_{min}$). Thus, the sampling probabilities of the whole layer can be improved.
The whole process can be defined as:
\begin{equation}
\tilde{p}_{min}= \begin{cases} p_{min}, & , \text { if } \varpi \geq 0 ; \\ p_{min} + \mu & , \text { otherwise }\end{cases}
\end{equation}
where $\mu$ represents the adjusting factor. Combining the sampling strategies on the spatial and temporal dimensions, we finally form our method, FP-Better.  FP-Better combines the sampling strategies on the spatial and
temporal dimensions to accelerate single-step adversarial training
via sampling lightweight subnetworks. The algorithm of FP-Better is summarized in  Algorithm~\ref{alg:FGSM_DSD}.
\par In this paper, the proposed FP-Better samples sub-networks by using dropping strategies, which are quite different from the previous Dropout~\cite{vivek2020single}. In detail,  the proposed FP-Better is different from the  Dropout~\cite{vivek2020single} in the following aspects. (1) In terms of motivation,  Dropout~\cite{vivek2020single} prevents catastrophic overfitting by randomly dropping some neurons during training. But the proposed method prevents catastrophic overfitting by randomly dropping some convolutional layers. Specifically, some residual blocks are randomly skipped and the shortcut path is kept. In this way, the proposed FP-Better is more efficient than  Dropout~\cite{vivek2020single}. (2) In terms of implementation, Dropout ~\cite{vivek2020single} only adopts a fixed dropping strategy. In this work, we propose a dynamic sampling strategy during training. (3) In terms of results, compared with Dropout~\cite{vivek2020single}, the proposed FP-Better achieves better adversarial robustness with less training time. {(Refer to Table~\ref{tb:more}. It an be observed that compared with the Dropout ~\cite{vivek2020single}, the proposed FB-Better achieves
the better robustness improvement under all adversarial attack
scenarios and the better training efficiency.)}

\begin{algorithm}[t]
\caption{FP-Better}
\label{alg:FGSM_DSD}
\begin{algorithmic}[1] 
\REQUIRE
The whole epoch $M$, the attack step size $\alpha$, the adversarial perturbation $\epsilon$, the label $y$, the clean image $\mathbf{x}$, the database size $N$, the whole model $f(\cdot)$, the sample model $f'(\cdot)$, the sampling probability parameters $p_{min}$, the adjusting factor $\mu$, and the network with parameters $\boldsymbol{\theta}$.
\STATE $\mathcal{L}_{\text{cur}}=0$
\STATE $\mathcal{L}_{\text{pre}}=0$
\FOR{$j=1,...,M$}
\IF{$\mathcal{L}_{\text{cur}} \geq \mathcal{L}_{\text{pre}}$} 
\STATE $\tilde{p}_{min}={p}_{min}$
\ELSE \STATE $\tilde{p}_{min}={p}_{min}+{\mu}$
\ENDIF
\STATE $\mathcal{L}_{\text{pre}} = \mathcal{L}_{\text{cur}}$
\STATE $ \boldsymbol{p}=[1,\tilde{p}_{min}]$

\FOR{$i=1,...,N$}
\STATE $f'_{i}(\cdot) \leftarrow \boldsymbol{p}$
\STATE $\boldsymbol{\varphi} =\mathbf{U}(-\epsilon ,\epsilon) $
\STATE $\nabla_{\text{grad}}=\nabla_{\mathbf{x}_{i}} \mathcal{L}(f'_i(\mathbf{x}_i+\boldsymbol{\varphi}, \boldsymbol{\theta}),y)$
\STATE $\boldsymbol{\delta}_{adv}=\Pi_{[-\epsilon, \epsilon]}\left[\varphi+\alpha \cdot \operatorname{sign}\left(\nabla_{\text{grad}}\right)\right]$
\STATE $\mathcal{L}_{\text{adv}}=\mathcal{L}(f(\mathbf{x}_{i} + \boldsymbol{\delta}_{adv},\boldsymbol{\theta}),y)$
\STATE $ \boldsymbol{\theta} \leftarrow \boldsymbol{\theta} -\nabla_{\boldsymbol{\theta}}\mathcal{L}_{\text{adv}} $
\STATE $\mathcal{L}_{\text{cur}}=\mathcal{L}_{\text{cur}}+\mathcal{L}_{\text{adv}}$
\ENDFOR
\ENDFOR
\end{algorithmic}
\end{algorithm}

\begin{table*}[]

\centering

\footnotesize

\caption{ The experiment of the hyper-parameter selection. Training time (minute), clean accuracy (\%) and robust accuracy (\%)   are reported  on  CIFAR-10 dataset using
ResNet18. Number in bold indicates the best.
}
\setlength\tabcolsep{0.38cm}
\begin{tabular}{@{}c|cc|cc|cc|cc|cc|cc@{}}
\toprule
\multirow{2}{*}{$\mu$} & \multicolumn{2}{c|}{0.1}                    & \multicolumn{2}{c|}{0.08}                   & \multicolumn{2}{c|}{0.06}          & \multicolumn{2}{c|}{0.04}                   & \multicolumn{2}{c|}{0.02}          & \multicolumn{2}{c}{0.01}          \\ \cmidrule(l){2-13} 
                   & \multicolumn{1}{c|}{Best}           & Last  & \multicolumn{1}{c|}{Best}  & Last           & \multicolumn{1}{c|}{Best}  & Last  & \multicolumn{1}{c|}{Best}           & Last  & \multicolumn{1}{c|}{Best}  & Last  & \multicolumn{1}{c|}{Best}  & Last  \\ \midrule \midrule
Clean              & \multicolumn{1}{c|}{83.88}          & 84.19 & \multicolumn{1}{c|}{84.06} & {84.07} & \multicolumn{1}{c|}{84.14} & \textbf{84.27} & \multicolumn{1}{c|}{83.98}          & 84.06 & \multicolumn{1}{c|}{83.35} & 83.47 & \multicolumn{1}{c|}{82.12} & 82.13 \\ \midrule
PGD-50             & \multicolumn{1}{c|}{\textbf{48.56}} & 48.25 & \multicolumn{1}{c|}{48.05} & 47.68          & \multicolumn{1}{c|}{48.27} & 47.73 & \multicolumn{1}{c|}{48.47}          & 47.96 & \multicolumn{1}{c|}{48.1}  & 47.56 & \multicolumn{1}{c|}{46.95} & 45.79 \\ \midrule
AA                 & \multicolumn{1}{c|}{45.19}          & 45.04 & \multicolumn{1}{c|}{43.97} & 43.95          & \multicolumn{1}{c|}{44.74} & 44.35 & \multicolumn{1}{c|}{\textbf{45.35}} & 44.84 & \multicolumn{1}{c|}{44.42} & 44.37 & \multicolumn{1}{c|}{43.75} & 43.06 \\ \midrule
Time(min)          & \multicolumn{2}{c|}{46}                     & \multicolumn{2}{c|}{46}                     & \multicolumn{2}{c|}{45}            & \multicolumn{2}{c|}{45}                     & \multicolumn{2}{c|}{44}            & \multicolumn{2}{c}{43}            \\ \bottomrule
\end{tabular}

\label{table:selection}
\end{table*}

\subsection{Relation to Catastrophic Overfitting} Catastrophic overfitting which leads to the failure of FGSM-AT~\cite{szegedy2013intriguing} is first noticed by Wong \emph{et al}.\cite{wong2020fast}. It refers to a phenomenon that during the training of single-step AT, the robustness accuracy against multi-step attack methods (such as PGD) suddenly decreases to 0\% after a few epochs. To overcome the overfitting,  Andriushchenko  \emph{et al}. \cite{andriushchenko2020understanding} and Kim \emph{et al}.\cite{ kim2020understanding} propose their own methods to prevent catastrophic overfitting from a different perspective. In detail, Andriushchenko \emph{et al}. \cite{andriushchenko2020understanding} propose to adopt a gradient regularization method to prevent the overfitting, called FGSM-GA. While Kim \emph{et al}.\cite{kim2020understanding} propose a simple yet effective method to adjust the attack step size for FGSM-AT during the training, called FGSM-CKPT. It  can also be regarded as a regularization method for the attack step size. However, they require more calculating time to perform their regularization methods. Specifically, FGSM-GA needs more calculating time to calculate model gradients to achieve regularization. FGSM-CKPT needs more calculating time to perform forwarding propagation to select the optimal attack step size for the generation of adversarial examples. Fortunately, based on Sec~\ref{Generalization_analysis}, the proposed FP-Better also has a regularizing effect. This kind of regularization can reduce the calculating time which is more efficient. 

\section{Experiments}
\label{Experiments}
To evaluate the effectiveness of the proposed FP-Better, we conduct extensive experiments on four benchmark image databases which are widely used to evaluate the adversarial robustness and training efficiency, \textit{i.e.,} CIFAR-10~\cite{krizhevsky2009learning}, CIFAR-100~\cite{krizhevsky2009learning}, Tiny ImageNet~\cite{deng2009imagenet}, and ImageNet~\cite{deng2009imagenet}. The CIFAR-10 consists of 50000 images in the training dataset and 10000 images in the testing dataset. It includes 10 classes with 
$32 \times 32$ image size. The CIFAR-100 also consists of 50000 images in the training dataset and  10000 images in the testing dataset.
It includes 100 classes with $32 \times 32$ image size. The Tiny ImageNet consists of 200 classes with 600 images in the
size of 64 × 64 for each class. ImageNet includes 1000 classes. The images in the ImageNet are resized to $224 \times 224 \times 3$. Following \cite{lee2020adversarial}, as for Tiny ImageNet and ImageNet, validation datasets are used to conduct comparative experiments. In this section, we first introduce the experimental settings which include the image datasets and experimental setups in Sec.~\ref{Experimental Settings}. 
We conduct a series of experiments to select the hyper-parameters used in our FP-Better in Sec.~\ref{Hyper-parameter}.
Then we compare the proposed  FP-Better with the previous single-step adversarial training methods in Sec.~\ref{Comparisons}. 
We conduct the ablation study to explore
the influence of each dimension on improving adversarial robustness in Sec.~\ref{Ablation}.

\begin{table*}[t]
\centering
\caption{Comparison results of training time (minute), clean accuracy (\%) and robust accuracy (\%) using ResNet18 on CIFAR-10 database using different adversarial training methods under $\ell_{\infty}= 8/225$. Number in bold indicates the best. 
}
\label{table:cifar10_ResNet}
\setlength\tabcolsep{0.5cm}
\begin{tabular}{ccccccccc}
\toprule
CIFAR10                         &      & Clean                           & PGD-10 & PGD-20         & PGD-50         & C\&W           & AA             & Time (min)            \\ \midrule
Standard Training      &  & \textbf{94.33}                         & 0.0  & 0.0          & 0.0        & 0.0          & 0.0         & 26                   \\ \midrule \midrule

\multirow{2}{*}{PGD-2-AT~\cite{rice2020overfitting}}       & Best & 86.84                           & 48.72  & 46.89          & 46.33          & 47.39          & 44.10          & \multirow{2}{*}{77}  \\ \cmidrule(lr){2-8}
                                & Last & 86.83                           & 48.21  & 46.6           & 46.19          & 47.05          & 43.81          &                      \\ \midrule 
\multirow{2}{*}{FGSM-RS~\cite{wong2020fast}}        & Best & 73.81                           & 42.31  & 41.55          & 41.26          & 39.84          & 37.07          & \multirow{2}{*}{51}  \\ \cmidrule(lr){2-8}
                                & Last & 83.82                           & 00.09  & 00.04          & 00.02          & 0.00           & 0.00           &                      \\ \midrule
\multirow{2}{*}{FGSM-CKPT~\cite{kim2020understanding}}      & Best & {90.29} & 41.96  & 39.84          & 39.15          & 41.13          & 37.15          & \multirow{2}{*}{76}  \\ \cmidrule(lr){2-8}
                                & Last & {90.29} & 41.96  & 39.84          & 39.15          & 41.13          & 37.15          &                      \\ \midrule
\multirow{2}{*}{FGSM-GA~\cite{andriushchenko2020understanding}  }      & Best & 83.96                           & 49.23  & 47.57          & 46.89          & 47.46          & 43.45          & \multirow{2}{*}{178} \\ \cmidrule(lr){2-8}
                                & Last & 84.43                           & 48.67  & 46.66          & 46.08          & 46.75          & 42.63          &                      \\ \midrule
\multirow{2}{*}{Free-AT(m=8)~\cite{shafahi2019adversarial}}   & Best & 80.38                           & 47.1   & 45.85          & 45.62          & 44.42          & 42.17          & \multirow{2}{*}{215} \\ \cmidrule(lr){2-8}
                                & Last & 80.75                           & 45.82  & 44.82          & 44.48          & 43.73          & 41.17          &                      \\ \midrule
\multirow{2}{*}{FP-Better (ours)} & Best & 83.98                           & \textbf{50.05}  & \textbf{48.76} & \textbf{48.47} & \textbf{48.09} & \textbf{45.35} & \multirow{2}{*}{45}  \\ \cmidrule(lr){2-8}
                                & Last & 84.06                           & \textbf{49.85}  & \textbf{48.31} & \textbf{47.96} & \textbf{47.62} & \textbf{44.84} &                      \\ \bottomrule
\end{tabular}

\end{table*}
\begin{table*}[h]
\centering
\caption{
Comparison results with YOPO of training time (minutes), clean accuracy (\%) and robust accuracy (\%) using ResNet18 on CIFAR-10 database under $\ell_{\infty}= 8/225$. Number in bold indicates the best.
}
\label{table:YOPO_v2}
\footnotesize
\setlength\tabcolsep{0.8cm}
\begin{tabular}{cccccc}
\toprule

Dataset                        & Method           & Clean          & PGD-50         & AA             & Time (min) \\ \midrule
\multirow{2}{*}{CIFAR-10}      & YOPO-5-3 ~\cite{zhang2019you}        & 83.99          & 42.93          & 40.38          & 118      \\ \cmidrule(l){2-6} 
                               & FB-Better (ours) & \textbf{84.06} & \textbf{47.96} & \textbf{44.84} & 45      \\ \midrule
\multirow{2}{*}{CIFAR-100}     & YOPO-5-3~\cite{zhang2019you}         & 57.44          & 20.69          & 18.31          & 148      \\ \cmidrule(l){2-6} 
                               & FB-Better (ours) & \textbf{59.42} & \textbf{28.23} & \textbf{23.76} & 57     \\ \midrule
\multirow{2}{*}{Tiny ImageNet} & YOPO-5-3 ~\cite{zhang2019you}        & 47.69          & 18.15          & 12.69          & 828     \\ \cmidrule(l){2-6} 
                               &FB-Better (ours) & \textbf{48.74} & \textbf{22.24} & \textbf{15.94} & 316     \\ \bottomrule

\end{tabular}
\end{table*}


\subsection{ Experimental Setups.} 
\label{Experimental Settings}
Following the setting of single-step AT methods \cite{wong2020fast,kim2020understanding,andriushchenko2020understanding}, as for CIFAR-10 and CIFAR-100, we adopt ResNet18~\cite{he2016deep} as the backbone network. As for Tiny ImageNet, we adopt PreActResNet18~\cite{he2016identity} as the backbone network. As for ImageNet, we make use of ResNet50~\cite{he2016deep} as the backbone network.
We adopt a  Stochastic Gradient Descent
(SGD) momentum optimizer with an initial learning rate of 0.1, the weight decay of $5 \times 10^{-4}$, and the momentum of 0.9. On CIFAR-10, CIFAR-100, and Tiny ImageNet, 
following the setting of ~\cite{rice2020overfitting, pang2020bag}, we set the total training epoch number to 110. And we adopt  a factor of 0.1 to decay the learning rate during the 100th and 105th epoch. On ImageNet, the total training epoch number is set to 90 following the setting of ~\cite{shafahi2019adversarial,wong2020fast}. And we adopt  a factor of 0.1 to decay the learning rate during the 30th and 60th epoch. As for our FP-Better, following the linear survival rate strategy, we set the initial survival rates to $[1,p_{min}]$ for all the blocks. The $p_{min}$ is set to 0.5. 
After adjusting survival rates, the survival rates are reset to $[1,p_{min}+\mu]$ where $\mu$ is a hyper-parameter. In this work, experiments of all AT methods are conducted on Tesla V100. We report the results of the last checkpoint and the results of the checkpoint with the best robust accuracy on the adversarial examples generated by PGD-10.
To comprehensively evaluate adversarial robustness, we adopt a series of attack methods which are widely used to evaluate adversarial robustness, including PGD~\cite{madry2017towards}, C\&W~\cite{carlini2017towards}, and autoattack (AA)~\cite{croce2020reliable} which consists of APGD-DLR~\cite{croce2020reliable}, APGD-CE~\cite{croce2020reliable}, FAB~\cite{croce2020minimally} and Square~\cite{andriushchenko2020square}.
Besides, the PGD attack method is conducted with 50, 20, and 10 iterations, called PGD-50, PGD-20, and PGD-10. One single NVIDIA Tesla V100 was used to conduct experiments. 
Moreover, we set the maximum perturbation strength $\epsilon$ to 8 under the $L_{\infty}$ to conduct evaluation experiments. There is a core hyper-parameter $\mu$ which controls the change
of survival rate in our FP-Better. We set $\mu$ to 0.04 to conduct comparison experiments. 
The selection of hyper-parameter $\mu$ is presented in Sec.~\ref{Hyper-parameter}.


\subsection{Detailed Hyper-parameter Settings}
\label{Hyper-parameter}
There is a core hyper-parameter $\mu$ which controls the change of survival rate in our FP-Better. It is not only related to adversarial robustness but also to training efficiency. We adopt ResNet18 on CIFAR-10 to conduct the experiment to select the hyper-parameter $\mu$ in the proposed method. The result is shown in Table~\ref{table:selection}. It can be observed that the training time of our FP-Better decreases
along with the increase of hyper-parameter $\mu$. When $\mu=0.04$, our FP-Better achieves the best adversarial robustness against AA which is a powerful attack method. Considering training efficiency, $\mu$ is set to 0.04 to conduct comparison experiments. Compared with FGSM-SD, consuming the same training time, the proposed FP-Better can achieve robustness performance under all attack scenarios. When training time is set to 45 minutes, under AA attack, our FP-Better achieves a higher robustness accuracy on the best and last checkpoints (45.35\% VS 43.34\%, 44.42\% VS 42.93\%).

\begin{figure}[t]
 \begin{center}
 \includegraphics[width=1\linewidth]{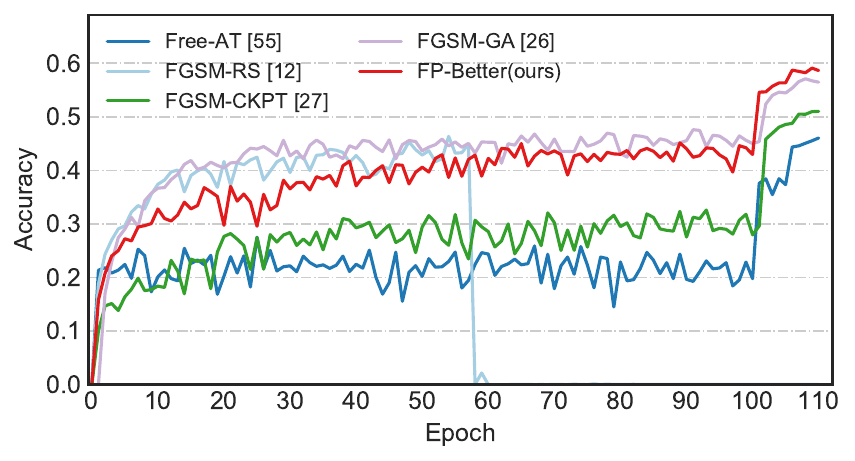}
\end{center}
  \caption{ The robust accuracy under PGD-10 attack of different single-step adversarial training methods on the training data of CIFAR-10 during the training phase.}
\label{fig:overfitting}
\end{figure}


\subsection{Comparisons with Previous Single-step Adversarial Training Methods}
\label{Comparisons}
We compare the proposed FP-Better with a series of previous single-step adversarial training methods which include FGSM-RS~\cite{wong2020fast}, FGSM-CKPT~\cite{kim2020understanding}, FGSM-GA~\cite{andriushchenko2020understanding}, and Free-AT~\cite{shafahi2019adversarial}. We also adopt a state-of-the-art multi-step adversarial training method (\textit{i.e.,} PGD-2-AT~\cite{rice2020overfitting} which makes use of two-step PGD for the adversarial example generation) as a powerful baseline. We adopt the optimal training hyper-parameters which are reported in the original works to conduct the adversarial training methods. Besides, to ensure comparison fairness, 
as for Free-AT, the epochs are not divided by $m$. The total epochs keep the same for these comparisons of adversarial training methods.

\begin{table}
\centering

\caption{More comparison results of training time (minute), clean accuracy (\%) and robust accuracy (\%) using ResNet18 on CIFAR-10 database under $\ell_{\infty}= 8/225$. Number in bold
indicates the best.}
\label{tb:more}
\setlength\tabcolsep{0.3cm}
\begin{tabular}{@{}c|c|c|c|c|c@{}}
\toprule
CIFAR-10        & Clean          & PGD-50         & C\&W           & AA             & Time(min) \\ \midrule
Stochastic~\cite{dhillon2018stochastic}      & 83.80          & 44.20          & 44.68          & 42.88          & 51        \\ \midrule
Ticket~\cite{li2020towards}          & 83.07          & 46.21          & 46.53          & 43.96          & 51        \\ \midrule
Dropout~\cite{vivek2020single}        & 82.01          & 45.08          & 45.21          & 43.17          & 51        \\ \midrule
FP-Better(ours) & \textbf{84.06} & \textbf{47.96} & \textbf{47.62} & \textbf{44.84} & 45        \\ \bottomrule
\end{tabular}
\end{table}
\begin{table*}[t]
\centering
\caption{Comparison results of training time (minute), clean accuracy (\%) and robust accuracy (\%) on  CIFAR-100 database using different adversarial training methods under $\ell_{\infty}= 8/225$. Number in bold indicates the best.
}
\label{table:cifar100_ResNet}

\setlength\tabcolsep{0.5cm}
\begin{tabular}{ccccccccc}
\toprule
CIFAR100                        &      & Clean          & PGD-10         & PGD-20         & PGD-50         & C\&W           & AA             & Time (min)           \\ \midrule
Standard Training      &  & \textbf{76.58}                         & 0.0  & 0.0          & 0.0        &0.0          & 0.0         & 35                  \\ \midrule \midrule

\multirow{2}{*}{PGD-2-AT~\cite{rice2020overfitting}}       & Best & 60.9           & 26.44          & 25.6           & 25.18          & 25.23          & 22.30          & \multirow{2}{*}{103}  \\ \cmidrule(lr){2-8}
                                & Last & \textbf{61.81} & 26.12          & 25.26          & 24.84          & 25.07          & 22.32          &                      \\ \midrule 
\multirow{2}{*}{FGSM-RS~\cite{wong2020fast}}        & Best & 49.85          & 22.47          & 22.01          & 21.82          & 20.55          & 18.29          & \multirow{2}{*}{70}  \\ \cmidrule(lr){2-8}
                                & Last & 60.55          & 00.45          & 00.25          & 00.19          & 00.25          & 0.00           &                      \\ \midrule
\multirow{2}{*}{FGSM-CKPT~\cite{kim2020understanding}}      & Best & \textbf{60.93} & 16.69          & 15.61          & 15.24          & 16.6           & 14.34          & \multirow{2}{*}{96}  \\ \cmidrule(lr){2-8}
                                & Last & 60.93          & 16.58          & 15.47          & 15.19          & 16.40          & 14.17          &                      \\ \midrule
\multirow{2}{*}{FGSM-GA~\cite{andriushchenko2020understanding}}        & Best & 54.35          & 22.93          & 22.36          & 22.2           & 21.2           & 18.88          & \multirow{2}{*}{187} \\ \cmidrule(lr){2-8}
                                & Last & 55.1           & 20.04          & 19.13          & 18.84          & 18.96          & 16.45          &                      \\ \midrule
\multirow{2}{*}{Free-AT(m=8)~\cite{shafahi2019adversarial}}   & Best & 52.49          & 24.07          & 23.52          & 23.36          & 21.66          & 19.47          & \multirow{2}{*}{229} \\ \cmidrule(lr){2-8}
                                & Last & 52.63          & 22.86          & 22.32          & 22.16          & 20.68          & 18.57          &                      \\ \midrule
\multirow{2}{*}{FP-Better(ours)} & Best & 59.05          & \textbf{29.51} & \textbf{28.64} & \textbf{28.35} & \textbf{26.71} & \textbf{23.76} & \multirow{2}{*}{57}  \\ \cmidrule(lr){2-8}
                                & Last & 59.42          & \textbf{29.2}  & \textbf{28.52} & \textbf{28.23} & \textbf{26.42} & \textbf{23.76} &                      \\ \bottomrule
\end{tabular}
\end{table*}
\subsubsection{Comparison Results on CIFAR-10} 
The comparison results of CIFAR-10 are shown in Table~\ref{table:cifar10_ResNet}. It can be observed that compared with other single-step adversarial training methods, our FP-Better not only achieves the best adversarial robustness under all adversarial attack scenarios but also achieves the highest training efficiency. In detail, under the PGD-50 attack, the previous single-step adversarial training models only achieve below 47\% robustness accuracy. Unlike them, our FP-Better can achieve more than 48\% robustness accuracy. Besides, under AA attack, the previous most robust single-step adversarial training method (FGSM-GA) achieves about 43\% robustness accuracy, while our FP-Better achieves about 45\% robustness accuracy. In terms of training efficiency, FGSM-RS which is the fastest training method in the previous AT methods requires 51 minutes to achieve training, while the proposed FP-Better only requires 45 minutes. Compared with the multi-step adversarial training method (PGD-2-AT) which makes use of an early stopping trick to improve adversarial robustness, the proposed FP-Better achieves better adversarial robustness under all attack scenarios. Moreover, the training process of the proposed FP-Better is about 1.7 times faster than PGD-2-AT. 
To investigate the effectiveness of our FP-Better, we also compare the proposed method with YOPO~\cite{zhang2019you} which also adopts the model with different layers to generate adversarial examples. The original YOPO uses the early stopping trick to improve adversarial robustness. To keep the performance of YOPO, we adopt the training settings to conduct YOPO. In this way, we adopt the training time of each epoch as the training efficiency metric. The result is shown in Table~\ref{table:YOPO_v2}.
It is clear that compared with YOPO, the proposed FB-Better can achieve higher clean and robust accuracy under all attacks on multiple scenarios. 

\par
\begin{figure}[t]
\begin{center}
   \includegraphics[width=0.95\linewidth]{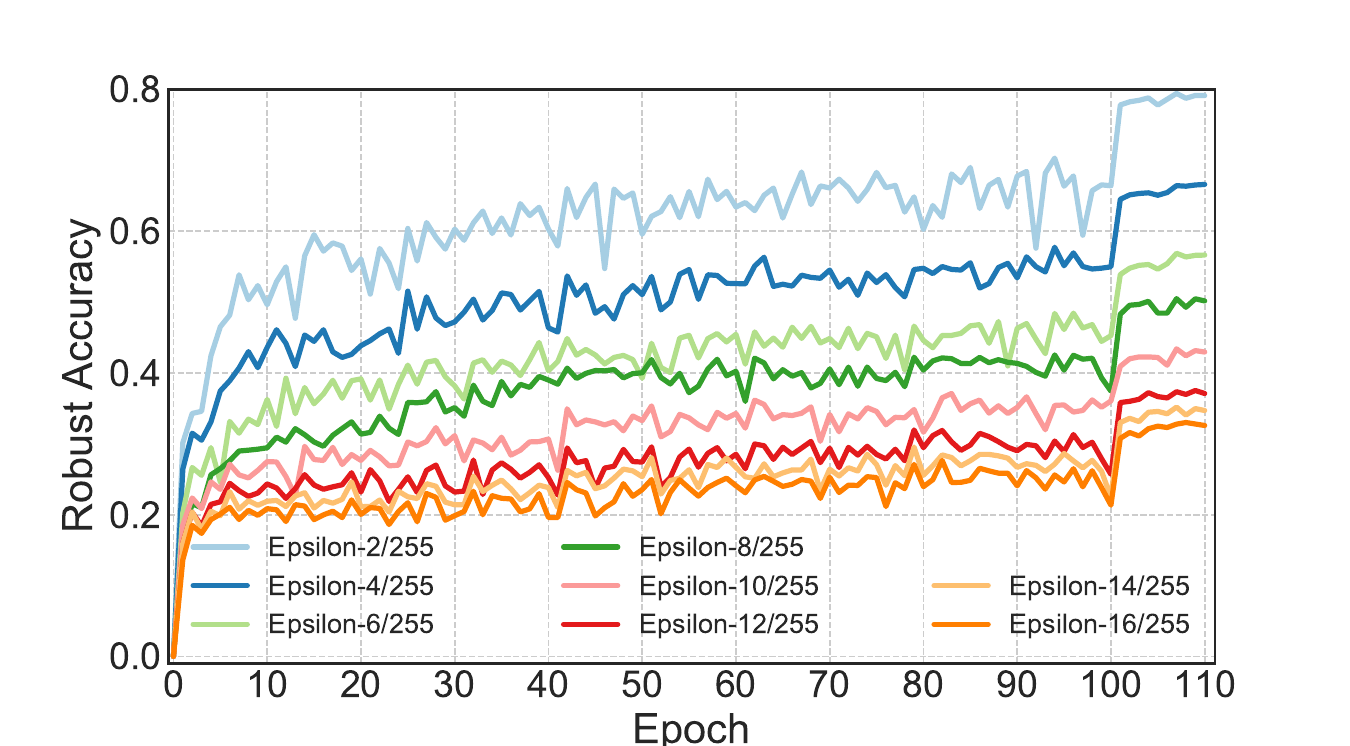}
\end{center}
\caption{ The robust accuracy of the proposed FB-Better under PGD-10 attack with the different attack strengths ($\ell_{\infty}= 2/225 \rightarrow 16/255$) for training and testing using ResNet18 on the CIFAR-10 during the training phase. 
}
\label{fig:attack_strength}
\end{figure}

\begin{figure*}
\begin{center}
 \includegraphics[width=1\linewidth]{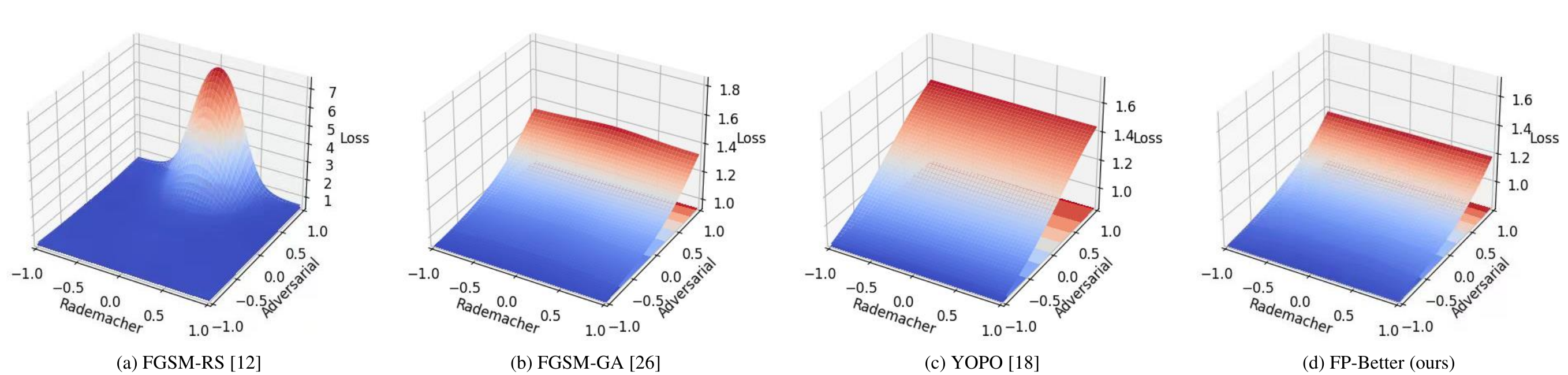}
\end{center}
\caption{ {Loss landscape of  FGSM-RS~\cite{wong2020fast}, YOPO~\cite{zhang2019you}, FGSM-GA~\cite{andriushchenko2020understanding}, and the proposed FP-Better on CIFAR-10. The loss landscape is generated by calculating the cross entropy loss on the space including a rademacher
direction and an adversarial direction. The rademacher direction is generated by a random perturbation. And the adversarial direction is generated by an adversarial perturbation of PGD-100.} }
\label{fig:loss_land}
\end{figure*}

 Catastrophic overfitting is one of the difficult problems for single-step adversarial training methods. To investigate the catastrophic overfitting, the robustness accuracy against PGD-10 is recorded during the training phase. 
Fig \ref{fig:overfitting} illustrates the robust accuracy curves under the attack of PGD-10.
It can be observed that the proposed FP-Better can prevent catastrophic overfitting like other advanced single-step adversaril training methods (FGSM-GA and FGSM-CKPT).  Compared with FGSM-GA and FGSM-CKPT, our FP-Better can achieve better robustness accuracy under the PGD-10 attack. Besides, following the default settings~\cite{andriushchenko2020understanding}, we adopt different attack strengths ($\ell_{\infty}= 2/225 \rightarrow 16/255$) for training and testing using ResNet18 on CIFAR-10. The robust accuracy evolution of the proposed FB-Better is shown in Fig.~\ref{fig:attack_strength}. It can be observed that under different attack strengths, the proposed FB-Better can also  prevent Catastrophic Overfitting. Besides following the previous work~\cite{DBLP:conf/nips/WuX020}, 
to study the loss landscape of the proposed FP-Better, we visualize the loss landscape
of the fast adversarial training models on CIFAR-10. In detail, the loss landscape is generated by calculating the cross entropy loss on the space including a rademacher direction and adversarial direction. The rademacher direction  is generated by a random perturbation. And the adversarial direction is generated by an
adversarial perturbation of PGD-100. As shown in Fig.~\ref{fig:loss_land}, it is clear that compared with other fast adversarial training methods, the
proposed FP-Better can achieve more linear cross-entropy loss in the adversarial direction, \emph{i.e.,} the proposed FP-Better can
achieve better adversarial robustness. 

\par We also compare the proposed method with other sampling-based adversarial training methods that include Stochastic~\cite{dhillon2018stochastic}, Ticket~\cite{li2020towards}, and Dropout~\cite{vivek2020single}. For a fair comparison, the same hyper-parameters that are used in our method
(see Sec.~\ref{Experimental Settings}) are adopted for them.  The result is
shown in Table~\ref{tb:more}. It can be observed that our FB-Better achieves
the best robustness improvement under all adversarial attack scenarios and the best training efficiency.
It indicates that the proposed method not only improves the adversarial robustness but also reduces the training time.

\subsubsection{Comparison Results on CIFAR-100} 
The comparison results of CIFAR-100 are shown in Table~\ref{table:cifar100_ResNet}. Similar phenomenons as on CIFAR-10 can be observed on CIFAR-100. Specifically, compared with other single-step AT methods, our FP-Better not only achieves the best adversarial robustness under all adversarial
attack scenarios but also achieves the highest training efficiency. For example, under the PGD-10 attack, the previous single-step AT models only achieve below 25\% robustness accuracy. But our FP-Better achieves about 29\% robustness accuracy which is even higher than the PGD-2-AT. In terms of training efficiency, the proposed method FP-Better is 1.2 times faster than FGSM-RS which is the fastest single-step AT method. 
Compared with the powerful PGD-2-AT, the proposed FP-Better consumes only 68\% of the training time of PGD-2-AT  yet achieves better adversarial robustness under all adversarial attack scenarios.
\begin{table}[t]
\centering
\caption{
Comparison results of  training time (minute), clean accuracy (\%) and robust accuracy (\%) using PreActResNet18 on the Tiny ImageNet database under $\ell_{\infty}= 8/225$. Number in bold indicates the best.
}
\setlength\tabcolsep{0.22cm}
\begin{tabular}{cccccc}
\toprule
Tiny ImageNet                   &      & Clean          & PGD-50         & AA             & Time (min)            \\ \midrule
Standard Training      &  & \textbf{56.73}                         & 0.0   & 0.0 & 169                   \\ \midrule \midrule
\multirow{2}{*}{PGD-2-AT~\cite{rice2020overfitting}}       & Best & 47.48          & 16.8           & 13.94          & \multirow{2}{*}{533}  \\ \cmidrule(lr){2-5}
                                & Last & 46.22          & 11.56          & 9.96           &                       \\ \midrule \midrule
\multirow{2}{*}{FGSM-RS~\cite{wong2020fast}}        & Best & 44.98          & 17.36          & 14.08          & \multirow{2}{*}{339}  \\ \cmidrule(lr){2-5}
                                & Last & 45.18          & 0.00           & 0.00           &                       \\ \midrule
\multirow{2}{*}{FGSM-CKPT~\cite{kim2020understanding}}      & Best & \textbf{49.98}          & 8.68           & 8.10           & \multirow{2}{*}{464}  \\ \cmidrule(lr){2-5}
                                & Last & \textbf{49.98}          & 8.68           & 8.10           &                       \\ \midrule
\multirow{2}{*}{FGSM-GA~\cite{andriushchenko2020understanding}}        & Best & 34.04          & 5.1            & 4.34           & \multirow{2}{*}{1054} \\ \cmidrule(lr){2-5}
                                & Last & 34.04          & 5.1            & 4.34           &                       \\ \midrule
\multirow{2}{*}{Free-AT(m=8)~\cite{shafahi2019adversarial}}   & Best & 38.9           & 11.02          & 9.28           & \multirow{2}{*}{1375} \\ \cmidrule(lr){2-5}
                                & Last & 40.06          & 8.2            & 7.34           &                       \\ \midrule
\multirow{2}{*}{FP-Better(ours)} & Best & 48.2  & \textbf{22.72} & \textbf{16.58} & \multirow{2}{*}{316}  \\ \cmidrule(lr){2-5}
                                & Last & 48.74 & \textbf{22.24} & \textbf{15.94} &                       \\ \bottomrule
\end{tabular}

\label{table:Tiny_Imagnet}
\end{table}
\begin{table}[]

\centering
\caption{Comparison results of  training time (hour), clean accuracy (\%) and robust accuracy (\%)) using ResNet50 on the ImageNet database under $\ell_{\infty}= 2/255, 4/255, 8/225$. Number in bold indicates the best.
}
\setlength\tabcolsep{0.15cm}

\begin{tabular}{c|c|c|c|c|c}
\toprule 
ImageNet                          & Epsilon      & Clean          & PGD-10         & PGD-50         & Time(hour)             \\\midrule \midrule
\multirow{3}{*}{Free-AT(m=4)~\cite{shafahi2019adversarial}}     & $\epsilon=$2 & 68.37          & 48.31          & 48.28          & \multirow{3}{*}{127.7} \\ 
                                  & $\epsilon=$4 & 63.42          & 33.22          & 33.08          &                        \\ 
                                  & $\epsilon=$8 & 52.09          & 19.46          & 12.92          &                        \\ \midrule
\multirow{3}{*}{FGSM-RS~\cite{wong2020fast}}          & $\epsilon=$2 & 67.65          & 48.78          & 48.67          & \multirow{3}{*}{44.5} \\ 
                                  & $\epsilon=$4 & 63.65          & 35.01          & 32.66          &                        \\ 
                                  & $\epsilon=$8 & \textbf{53.89}          & 0.00           & 0.00           &                        \\ \midrule
\multirow{3}{*}{FP-Better(ours)} & $\epsilon=$2 & \textbf{68.44} & \textbf{49.01} & \textbf{48.90} & \multirow{3}{*}{40.2} \\ 
                                  & $\epsilon=$4 & \textbf{64.52} & \textbf{36.04} & \textbf{33.63} &                        \\ 
                                  & $\epsilon=$8 & 52.96 & \textbf{20.86} & \textbf{13.43} &                        \\ \bottomrule
\end{tabular}

\label{table:Imagnet}
\end{table}
\subsubsection{Comparison Results on Tiny ImageNet} 
Compared with the previous images databases, Tiny ImageNet contains more images with larger size, which makes it harder to obtain adversarial robustness on it.  
The comparison results of Tiny ImageNet are shown in Table~\ref{table:Tiny_Imagnet}. It can be observed that compared with competing single-step adversarial training methods, our FP-Better has higher robust accuracy and training efficiency. Compared with the  powerful PGD-2-AT, our FP-Better also achieves better adversarial robustness. For example, under AA attack, PGD-2-AT achieves robust performance of about 14\% and 10\% robustness accuracy at the best and last checkpoints, but the proposed FP-Better achieves
the robust performance of about 16\% and 15\% robustness accuracy, respectively. Note that the proposed FP-Better is about 1.7 times faster than PGD-2-AT. More importantly, our FP-Better is about 1.1 times faster than FGSM-RS which is the fastest adversarial training method of the previous.

\subsubsection{Comparison Results on ImageNet} 
ImageNet is a large image dataset which is widely used for image classification. {Compared with previous datasets, the ImageNet covers more images and classes. It is hard for the image classification model to achieve adversarial robustness on ImageNet.}
Conducting adversarial training on it requires more training costs. 
Following ~\cite{shafahi2019adversarial,wong2020fast}, we set the maximum perturbation
strength $\epsilon$ to 2/255, 4/255, and 8/255 to conduct comparison experiments using ResNet50.
The comparison results of ImageNet are shown in Table~\ref{table:Imagnet}. We can observe that when $\epsilon=2/255$ and $\epsilon=4/255$, all adversarial training methods can achieve the same adversarial robustness. They achieve the performance of about 48\%  robustness accuracy against PGD-10 and PGD-50 attacks. However, when the maximum perturbation
strength becomes larger, the proposed FP-Better can achieve the better adversarial robustness against PGD attack. In detail, when $\epsilon=8/255$, Free-AT achieves
about 19\% and 12\% robustness accuracy under the attacks of PGD-10 and PGD-50. But our FP-Better achieves about 13\% and 21\% robust accuracy. More importantly, our FP-Better can be 3.2 times faster than Free-AT.
\begin{table}[t]
\centering
\caption{
Ablation study of the our FP-Better. Comparison results of  training time (minute), clean accuracy (\%) and robust accuracy (\%) using ResNet18 on the CIFAR-10 database under $\ell_{\infty}= 8/225$. Number in bold indicates the best. 
}
\label{table:ablation}
\setlength\tabcolsep{0.2cm}
\begin{tabular}{@{}c|c|c|c|c|c@{}}
\toprule
Sampling strategy                   &      & Clean          & PGD-50         & AA             & Time (min)          \\ \midrule
\multirow{2}{*}{Spatial dimension}  & Best & 78.22          & 44.02          & 41.39          & \multirow{2}{*}{41} \\ \cmidrule(lr){2-5}
                                    & Last & 80.42          & 43.39          & 40.64          &                     \\ \midrule
\multirow{2}{*}{Temporal dimension} & Best & 83.21          & 47.52          & 44.22          & \multirow{2}{*}{41} \\ \cmidrule(lr){2-5}
                                    & Last &         83.21          & 47.52          & 44.22              &                     \\ \midrule
\multirow{2}{*}{Both}               & Best & \textbf{83.98} & \textbf{48.47} & \textbf{45.35} & \multirow{2}{*}{45} \\ \cmidrule(lr){2-5}
                                    & Last & \textbf{84.06} & \textbf{47.96} & \textbf{44.84} &                     \\ \bottomrule
\end{tabular}
\end{table}
\subsection{Ablation Study}
\label{Ablation}
In this paper, we propose a novel sampling strategy from the spatial and temporal dimensions. To  study
the influence of each dimension on improving the adversarial robustness, we conduct an ablation study on the CIFAR-10
dataset using ResNet18. In detail, as for only using the 
sampling strategy on the spatial dimension, we set the linear decaying survival probability to $[1,0.5]$ and the adjusting factor $\mu$ to 0. 
As for only using the 
sampling strategy on the temporal dimension, we adopt the uniform survival probability which is set to 0.5 for the while layers, the adjusting factor $\mu$ to 0.04. 
The result is shown in Table~\ref{table:ablation}. It can be observed that compared with the sampling strategy on the spatial dimension, the sampling strategy on the temporal dimensions can achieve better adversarial robustness under the PGD-50 and AA attacks. It is more important to dynamically adjust the sampling strategy over time. Combining the sampling strategies on the spatial and temporal dimension, the proposed FP-Better can achieve the best clean and robust performance.  Compared with the sampling strategy in the spatial dimension \cite{huang2016deep}, our sampling strategy in the temporal dimension is more suitable for adversarial training to improve adversarial robustness.

\section{Conclusion}
\label{sec:conclusion}
 We accelerate the single-step adversarial training by sampling subnetwork from the whole network to conduct adversarial training.
By doing this, both the forward and backward passes can be accelerated. We propose a novel sampling strategy to sample subnetworks from both temporal and spatial dimensions. The sampling varies from layer to layer and from iteration to iteration. Compared with previous single-step adversarial training methods, we not only achieve better model robustness but also reduce the training cost. Evaluations on four image databases demonstrate that the proposed FB-Better prevents catastrophic overfitting and outperforms state-of-the-art single-step adversarial training methods. Our code has been released at https://github.com/jiaxiaojunQAQ/FP-Better.

\appendix
\label{sec:proof}

This appendix gives the proof for Theorem 3.1. The proof is based on He \emph{et.al}. \cite{he2020robustness}. We recall some of the paper to make this paper completed.

We first define the following term to measure the intensity of adversarial learning.

\begin{definition}[Robustified Intensity]
\label{def:ri}
For adversarial training, the robustified intensity is defined to be
\begin{equation}
\label{eq:ri}
I = \frac{ \max_{\theta, x, y} \left \| \nabla_{\theta} \max_{\Vert x^\prime - x \Vert \leq \rho} l (h_{\theta} (x^\prime), y) \right \|}{\max_{\theta, x, y} \left \| \nabla_{\theta} l (h_{\theta} (x), y) \right \|},
\end{equation}
where $\| \cdot \|$ is a norm defined in the space of the gradient.
\end{definition}

Empirical study shows that the gradient noise satisfies the Laplacian assumption as follows.

\begin{assumption}
The gradient calculated from a mini-batch is drawn from a Laplacian distribution centered at the empirical risk,
\begin{align*}
\frac{1}{\tau} \sum_{(x, y) \in \mathcal B} \nabla_\theta \max_{\Vert x^\prime - x \Vert \leq \rho} l (h_\theta( x^\prime), y) \sim \mathrm{Lap} \left(\nabla_\theta \hat{\mathcal R}_S^A (\theta), b\right).
\end{align*}
\end{assumption}

Then, we have the following theorem under Laplacian assumption.

\begin{theorem}
\label{thm:privacy_whole}
Suppose one employs SGD for adversarial training. $L_{ERM}$ is the maximal gradient norm in ERM. Also, suppose the whole training procedure has $T$ iterations.
Then, the adversarial training is $(\varepsilon, \delta)$-differentially private, where
\begin{gather*}
	\label{eq:epsilon_whole}
	\varepsilon = \varepsilon_0\sqrt{2 T \log \frac{N}{\delta'}} +T \varepsilon_0 (e^{\varepsilon_0} - 1),\\
	\delta =  \frac{\delta'}{N},~
	\end{gather*}
	in which
	\begin{gather*}
	\varepsilon_0= \frac{2 L_{ERM}}{N b} I,
	\end{gather*}
and $\delta'$ is a positive real, $\tau$ is the batch size, $I$ is the robustified intensity, and $b$ is the Laplace parameter.
\end{theorem}

Recall the following theorem from \cite{he2020robustness}.

\begin{theorem}[High-Probability Generalization Bound via Differential Privacy]
\label{thm:high_probability_privacy}
Suppose all conditions of Theorem \ref{thm:privacy_whole} hold. Then, the algorithm $\mathcal A$ has a high-probability generalization bound as follows. Specifically, the following inequality holds with probability at least $1 - \gamma$:
\begin{align}
\label{eq:high_probability_privacy}
   & \mathbb{E}_{\mathcal{A}}\mathcal R(\mathcal{A}(S)) - \mathbb{E}_{\mathcal{A}}\hat{\mathcal R}_S(\mathcal{A}(S)) \nonumber\\
   \le & c \left(M(1-e^{-\varepsilon}+e^{-\varepsilon}\delta) \log{N}\log{\frac{N}{\gamma}}+\sqrt{\frac{\log 1 / \gamma}{ N}} \right),
\end{align}
where $\gamma$ is an arbitrary probability mass, $M$ is the bound for loss $l$, $N$ is the training sample size, $c$ is a universal constant for any sample distribution,  and the probability is defined over the sample set $S$.
\end{theorem}

Then, we may prove Theorem 3.1.
\bibliographystyle{IEEEtran}
\bibliography{refer}

\begin{thebibliography}{10}
\providecommand{\url}[1]{#1}
\csname url@samestyle\endcsname
\providecommand{\newblock}{\relax}
\providecommand{\bibinfo}[2]{#2}
\providecommand{\BIBentrySTDinterwordspacing}{\spaceskip=0pt\relax}
\providecommand{\BIBentryALTinterwordstretchfactor}{4}
\providecommand{\BIBentryALTinterwordspacing}{\spaceskip=\fontdimen2\font plus
\BIBentryALTinterwordstretchfactor\fontdimen3\font minus
  \fontdimen4\font\relax}
\providecommand{\BIBforeignlanguage}[2]{{%
\expandafter\ifx\csname l@#1\endcsname\relax
\typeout{** WARNING: IEEEtran.bst: No hyphenation pattern has been}%
\typeout{** loaded for the language `#1'. Using the pattern for}%
\typeout{** the default language instead.}%
\else
\language=\csname l@#1\endcsname
\fi
#2}}
\providecommand{\BIBdecl}{\relax}
\BIBdecl

\bibitem{szegedy2013intriguing}
C.~Szegedy, W.~Zaremba, I.~Sutskever, J.~Bruna, D.~Erhan, I.~J. Goodfellow, and
  R.~Fergus, ``Intriguing properties of neural networks,'' in \emph{2nd
  International Conference on Learning Representations, {ICLR} 2014, Banff, AB,
  Canada, April 14-16, 2014, Conference Track Proceedings}, Y.~Bengio and
  Y.~LeCun, Eds., 2014.

\bibitem{goodfellow2014explaining}
I.~J. Goodfellow, J.~Shlens, and C.~Szegedy, ``Explaining and harnessing
  adversarial examples,'' in \emph{3rd International Conference on Learning
  Representations, {ICLR} 2015, San Diego, CA, USA, May 7-9, 2015, Conference
  Track Proceedings}, Y.~Bengio and Y.~LeCun, Eds., 2015.

\bibitem{li2019nattack}
Y.~Li, L.~Li, L.~Wang, T.~Zhang, and B.~Gong, ``{NATTACK:} learning the
  distributions of adversarial examples for an improved black-box attack on
  deep neural networks,'' in \emph{Proceedings of the 36th International
  Conference on Machine Learning, {ICML} 2019, 9-15 June 2019, Long Beach,
  California, {USA}}, ser. Proceedings of Machine Learning Research,
  K.~Chaudhuri and R.~Salakhutdinov, Eds., vol.~97.\hskip 1em plus 0.5em minus
  0.4em\relax {PMLR}, 2019, pp. 3866--3876.

\bibitem{dong2018boosting}
Y.~Dong, F.~Liao, T.~Pang, H.~Su, J.~Zhu, X.~Hu, and J.~Li, ``Boosting
  adversarial attacks with momentum,'' in \emph{2018 {IEEE} Conference on
  Computer Vision and Pattern Recognition, {CVPR} 2018, Salt Lake City, UT,
  USA, June 18-22, 2018}.\hskip 1em plus 0.5em minus 0.4em\relax Computer
  Vision Foundation / {IEEE} Computer Society, 2018, pp. 9185--9193.

\bibitem{wang2021enhancing}
X.~Wang and K.~He, ``Enhancing the transferability of adversarial attacks
  through variance tuning,'' in \emph{{IEEE} Conference on Computer Vision and
  Pattern Recognition, {CVPR} 2021, virtual, June 19-25, 2021}.\hskip 1em plus
  0.5em minus 0.4em\relax Computer Vision Foundation / {IEEE}, 2021, pp.
  1924--1933.

\bibitem{bai2021targeted}
J.~Bai, B.~Wu, Y.~Zhang, Y.~Li, Z.~Li, and S.~Xia, ``Targeted attack against
  deep neural networks via flipping limited weight bits,'' in \emph{9th
  International Conference on Learning Representations, {ICLR} 2021, Virtual
  Event, Austria, May 3-7, 2021}.\hskip 1em plus 0.5em minus 0.4em\relax
  OpenReview.net, 2021.

\bibitem{fan2020sparse}
Y.~Fan, B.~Wu, T.~Li, Y.~Zhang, M.~Li, Z.~Li, and Y.~Yang, ``Sparse adversarial
  attack via perturbation factorization,'' in \emph{Computer Vision - {ECCV}
  2020 - 16th European Conference, Glasgow, UK, August 23-28, 2020,
  Proceedings, Part {XXII}}, ser. Lecture Notes in Computer Science,
  A.~Vedaldi, H.~Bischof, T.~Brox, and J.~Frahm, Eds., vol. 12367.\hskip 1em
  plus 0.5em minus 0.4em\relax Springer, 2020, pp. 35--50.

\bibitem{liu2022watermark}
X.~Liu, J.~Liu, Y.~Bai, J.~Gu, T.~Chen, X.~Jia, and X.~Cao, ``Watermark
  vaccine: Adversarial attacks to prevent watermark removal,'' in
  \emph{Computer Vision - {ECCV} 2022 - 17th European Conference, Tel Aviv,
  Israel, October 23-27, 2022, Proceedings, Part {XIV}}, ser. Lecture Notes in
  Computer Science, S.~Avidan, G.~J. Brostow, M.~Ciss{\'{e}}, G.~M. Farinella,
  and T.~Hassner, Eds., vol. 13674.\hskip 1em plus 0.5em minus 0.4em\relax
  Springer, 2022, pp. 1--17.

\bibitem{liang2022large}
S.~Liang, L.~Li, Y.~Fan, X.~Jia, J.~Li, B.~Wu, and X.~Cao, ``A large-scale
  multiple-objective method for black-box attack against object detection,'' in
  \emph{Computer Vision - {ECCV} 2022 - 17th European Conference, Tel Aviv,
  Israel, October 23-27, 2022, Proceedings, Part {IV}}, ser. Lecture Notes in
  Computer Science, S.~Avidan, G.~J. Brostow, M.~Ciss{\'{e}}, G.~M. Farinella,
  and T.~Hassner, Eds., vol. 13664.\hskip 1em plus 0.5em minus 0.4em\relax
  Springer, 2022, pp. 619--636.

\bibitem{athalye2018obfuscated}
A.~Athalye, N.~Carlini, and D.~A. Wagner, ``Obfuscated gradients give a false
  sense of security: Circumventing defenses to adversarial examples,'' in
  \emph{Proceedings of the 35th International Conference on Machine Learning,
  {ICML} 2018, Stockholmsm{\"{a}}ssan, Stockholm, Sweden, July 10-15, 2018},
  ser. Proceedings of Machine Learning Research, J.~G. Dy and A.~Krause, Eds.,
  vol.~80.\hskip 1em plus 0.5em minus 0.4em\relax {PMLR}, 2018, pp. 274--283.

\bibitem{madry2017towards}
A.~Madry, A.~Makelov, L.~Schmidt, D.~Tsipras, and A.~Vladu, ``Towards deep
  learning models resistant to adversarial attacks,'' in \emph{6th
  International Conference on Learning Representations, {ICLR} 2018, Vancouver,
  BC, Canada, April 30 - May 3, 2018, Conference Track Proceedings}.\hskip 1em
  plus 0.5em minus 0.4em\relax OpenReview.net, 2018.

\bibitem{wong2020fast}
E.~Wong, L.~Rice, and J.~Z. Kolter, ``Fast is better than free: Revisiting
  adversarial training,'' in \emph{8th International Conference on Learning
  Representations, {ICLR} 2020, Addis Ababa, Ethiopia, April 26-30,
  2020}.\hskip 1em plus 0.5em minus 0.4em\relax OpenReview.net, 2020.

\bibitem{DBLP:journals/tifs/HuangJGLP22}
Y.~Huang, F.~Juefei{-}Xu, Q.~Guo, Y.~Liu, and G.~Pu, ``Fakelocator: Robust
  localization of gan-based face manipulations,'' \emph{{IEEE} Trans. Inf.
  Forensics Secur.}, vol.~17, pp. 2657--2672, 2022.

\bibitem{li2022semi}
Y.~Li, B.~Wu, Y.~Feng, Y.~Fan, Y.~Jiang, Z.~Li, and S.~Xia, ``Semi-supervised
  robust training with generalized perturbed neighborhood,'' \emph{Pattern
  Recognit.}, vol. 124, p. 108472, 2022.

\bibitem{jia2022adversarial}
X.~Jia, Y.~Zhang, B.~Wu, K.~Ma, J.~Wang, and X.~Cao, ``{LAS-AT:} adversarial
  training with learnable attack strategy,'' in \emph{{IEEE/CVF} Conference on
  Computer Vision and Pattern Recognition, {CVPR} 2022, New Orleans, LA, USA,
  June 18-24, 2022}.\hskip 1em plus 0.5em minus 0.4em\relax {IEEE}, 2022, pp.
  13\,388--13\,398.

\bibitem{DBLP:conf/nips/MaoCDZQYLZ022}
X.~Mao, Y.~Chen, R.~Duan, Y.~Zhu, G.~Qi, S.~Ye, X.~Li, R.~Zhang, and H.~Xue,
  ``Enhance the visual representation via discrete adversarial training,'' in
  \emph{NeurIPS}, 2022.

\bibitem{mao2022towards}
X.~Mao, G.~Qi, Y.~Chen, X.~Li, R.~Duan, S.~Ye, Y.~He, and H.~Xue, ``Towards
  robust vision transformer,'' in \emph{{IEEE/CVF} Conference on Computer
  Vision and Pattern Recognition, {CVPR} 2022, New Orleans, LA, USA, June
  18-24, 2022}.\hskip 1em plus 0.5em minus 0.4em\relax {IEEE}, 2022, pp.
  12\,032--12\,041.

\bibitem{zhang2019you}
D.~Zhang, T.~Zhang, Y.~Lu, Z.~Zhu, and B.~Dong, ``You only propagate once:
  Accelerating adversarial training via maximal principle,'' in \emph{Advances
  in Neural Information Processing Systems 32: Annual Conference on Neural
  Information Processing Systems 2019, NeurIPS 2019, December 8-14, 2019,
  Vancouver, BC, Canada}, H.~M. Wallach, H.~Larochelle, A.~Beygelzimer,
  F.~d'Alch{\'{e}}{-}Buc, E.~B. Fox, and R.~Garnett, Eds., 2019, pp. 227--238.

\bibitem{DBLP:conf/mm/00010JMMLP21}
Y.~Huang, Q.~Guo, F.~Juefei{-}Xu, L.~Ma, W.~Miao, Y.~Liu, and G.~Pu,
  ``Advfilter: Predictive perturbation-aware filtering against adversarial
  attack via multi-domain learning,'' in \emph{{MM} '21: {ACM} Multimedia
  Conference, Virtual Event, China, October 20 - 24, 2021}, H.~T. Shen,
  Y.~Zhuang, J.~R. Smith, Y.~Yang, P.~C{\'{e}}sar, F.~Metze, and
  B.~Prabhakaran, Eds.\hskip 1em plus 0.5em minus 0.4em\relax {ACM}, 2021, pp.
  395--403.

\bibitem{dai2021parameterizing}
S.~Dai, S.~Mahloujifar, and P.~Mittal, ``Parameterizing activation functions
  for adversarial robustness,'' in \emph{43rd {IEEE} Security and Privacy, {SP}
  Workshops 2022, San Francisco, CA, USA, May 22-26, 2022}.\hskip 1em plus
  0.5em minus 0.4em\relax {IEEE}, 2022, pp. 80--87.

\bibitem{bai2021improving}
Y.~Bai, Y.~Zeng, Y.~Jiang, S.~Xia, X.~Ma, and Y.~Wang, ``Improving adversarial
  robustness via channel-wise activation suppressing,'' in \emph{9th
  International Conference on Learning Representations, {ICLR} 2021, Virtual
  Event, Austria, May 3-7, 2021}.\hskip 1em plus 0.5em minus 0.4em\relax
  OpenReview.net, 2021.

\bibitem{cui2021learnable}
J.~Cui, S.~Liu, L.~Wang, and J.~Jia, ``Learnable boundary guided adversarial
  training,'' in \emph{2021 {IEEE/CVF} International Conference on Computer
  Vision, {ICCV} 2021, Montreal, QC, Canada, October 10-17, 2021}.\hskip 1em
  plus 0.5em minus 0.4em\relax {IEEE}, 2021, pp. 15\,701--15\,710.

\bibitem{DBLP:conf/nips/ZieglerNCBSLSNW22}
D.~M. Ziegler, S.~Nix, L.~Chan, T.~Bauman, P.~Schmidt{-}Nielsen, T.~Lin,
  A.~Scherlis, N.~Nabeshima, B.~Weinstein{-}Raun, D.~de~Haas, B.~Shlegeris, and
  N.~Thomas, ``Adversarial training for high-stakes reliability,'' in
  \emph{NeurIPS}, 2022.

\bibitem{duan2021advdrop}
R.~Duan, Y.~Chen, D.~Niu, Y.~Yang, A.~K. Qin, and Y.~He, ``Advdrop: Adversarial
  attack to dnns by dropping information,'' in \emph{2021 {IEEE/CVF}
  International Conference on Computer Vision, {ICCV} 2021, Montreal, QC,
  Canada, October 10-17, 2021}.\hskip 1em plus 0.5em minus 0.4em\relax {IEEE},
  2021, pp. 7486--7495.

\bibitem{duan2021adversarial}
R.~Duan, X.~Mao, A.~K. Qin, Y.~Chen, S.~Ye, Y.~He, and Y.~Yang, ``Adversarial
  laser beam: Effective physical-world attack to dnns in a blink,'' in
  \emph{{IEEE} Conference on Computer Vision and Pattern Recognition, {CVPR}
  2021, virtual, June 19-25, 2021}.\hskip 1em plus 0.5em minus 0.4em\relax
  Computer Vision Foundation / {IEEE}, 2021, pp. 16\,062--16\,071.

\bibitem{andriushchenko2020understanding}
M.~Andriushchenko and N.~Flammarion, ``Understanding and improving fast
  adversarial training,'' in \emph{Advances in Neural Information Processing
  Systems 33: Annual Conference on Neural Information Processing Systems 2020,
  NeurIPS 2020, December 6-12, 2020, virtual}, H.~Larochelle, M.~Ranzato,
  R.~Hadsell, M.~Balcan, and H.~Lin, Eds., 2020.

\bibitem{kim2020understanding}
H.~Kim, W.~Lee, and J.~Lee, ``Understanding catastrophic overfitting in
  single-step adversarial training,'' in \emph{Thirty-Fifth {AAAI} Conference
  on Artificial Intelligence, {AAAI} 2021, Thirty-Third Conference on
  Innovative Applications of Artificial Intelligence, {IAAI} 2021, The Eleventh
  Symposium on Educational Advances in Artificial Intelligence, {EAAI} 2021,
  Virtual Event, February 2-9, 2021}.\hskip 1em plus 0.5em minus 0.4em\relax
  {AAAI} Press, 2021, pp. 8119--8127.

\bibitem{sriramanan2020guided}
G.~Sriramanan, S.~Addepalli, A.~Baburaj, and V.~B. R., ``Guided adversarial
  attack for evaluating and enhancing adversarial defenses,'' in \emph{Advances
  in Neural Information Processing Systems 33: Annual Conference on Neural
  Information Processing Systems 2020, NeurIPS 2020, December 6-12, 2020,
  virtual}, H.~Larochelle, M.~Ranzato, R.~Hadsell, M.~Balcan, and H.~Lin, Eds.,
  2020.

\bibitem{sriramanan2021towards}
------, ``Towards efficient and effective adversarial training,'' in
  \emph{Advances in Neural Information Processing Systems 34: Annual Conference
  on Neural Information Processing Systems 2021, NeurIPS 2021, December 6-14,
  2021, virtual}, 2021, pp. 11\,821--11\,833.

\bibitem{xiong2022stochastic}
Y.~Xiong, J.~Lin, M.~Zhang, J.~E. Hopcroft, and K.~He, ``Stochastic variance
  reduced ensemble adversarial attack for boosting the adversarial
  transferability,'' in \emph{{IEEE/CVF} Conference on Computer Vision and
  Pattern Recognition, {CVPR} 2022, New Orleans, LA, USA, June 18-24,
  2022}.\hskip 1em plus 0.5em minus 0.4em\relax {IEEE}, 2022, pp.
  14\,963--14\,972.

\bibitem{yuan2022adaptive}
Z.~Yuan, J.~Zhang, and S.~Shan, ``Adaptive image transformations for
  transfer-based adversarial attack,'' in \emph{Computer Vision - {ECCV} 2022 -
  17th European Conference, Tel Aviv, Israel, October 23-27, 2022, Proceedings,
  Part {V}}, ser. Lecture Notes in Computer Science, S.~Avidan, G.~J. Brostow,
  M.~Ciss{\'{e}}, G.~M. Farinella, and T.~Hassner, Eds., vol. 13665.\hskip 1em
  plus 0.5em minus 0.4em\relax Springer, 2022, pp. 1--17.

\bibitem{zhu2022toward}
Y.~Zhu, Y.~Chen, X.~Li, K.~Chen, Y.~He, X.~Tian, B.~Zheng, Y.~Chen, and
  Q.~Huang, ``Toward understanding and boosting adversarial transferability
  from a distribution perspective,'' \emph{{IEEE} Trans. Image Process.},
  vol.~31, pp. 6487--6501, 2022.

\bibitem{li2022subspace}
T.~Li, Y.~Wu, S.~Chen, K.~Fang, and X.~Huang, ``Subspace adversarial
  training,'' in \emph{{IEEE/CVF} Conference on Computer Vision and Pattern
  Recognition, {CVPR} 2022, New Orleans, LA, USA, June 18-24, 2022}.\hskip 1em
  plus 0.5em minus 0.4em\relax {IEEE}, 2022, pp. 13\,399--13\,408.

\bibitem{jia2022boosting}
X.~Jia, Y.~Zhang, B.~Wu, J.~Wang, and X.~Cao, ``Boosting fast adversarial
  training with learnable adversarial initialization,'' \emph{{IEEE} Trans.
  Image Process.}, vol.~31, pp. 4417--4430, 2022.

\bibitem{vivek2020single}
V.~B. S. and R.~V. Babu, ``Single-step adversarial training with dropout
  scheduling,'' in \emph{2020 {IEEE/CVF} Conference on Computer Vision and
  Pattern Recognition, {CVPR} 2020, Seattle, WA, USA, June 13-19, 2020}.\hskip
  1em plus 0.5em minus 0.4em\relax Computer Vision Foundation / {IEEE}, 2020,
  pp. 947--956.

\bibitem{jia2022prior}
X.~Jia, Y.~Zhang, X.~Wei, B.~Wu, K.~Ma, J.~Wang, and X.~Cao, ``Prior-guided
  adversarial initialization for fast adversarial training,'' in \emph{Computer
  Vision - {ECCV} 2022 - 17th European Conference, Tel Aviv, Israel, October
  23-27, 2022, Proceedings, Part {IV}}, ser. Lecture Notes in Computer Science,
  S.~Avidan, G.~J. Brostow, M.~Ciss{\'{e}}, G.~M. Farinella, and T.~Hassner,
  Eds., vol. 13664.\hskip 1em plus 0.5em minus 0.4em\relax Springer, 2022, pp.
  567--584.

\bibitem{moosavi2016deepfool}
S.~Moosavi{-}Dezfooli, A.~Fawzi, and P.~Frossard, ``Deepfool: {A} simple and
  accurate method to fool deep neural networks,'' in \emph{2016 {IEEE}
  Conference on Computer Vision and Pattern Recognition, {CVPR} 2016, Las
  Vegas, NV, USA, June 27-30, 2016}.\hskip 1em plus 0.5em minus 0.4em\relax
  {IEEE} Computer Society, 2016, pp. 2574--2582.

\bibitem{tramer2017ensemble}
F.~Tram{\`{e}}r, A.~Kurakin, N.~Papernot, I.~J. Goodfellow, D.~Boneh, and P.~D.
  McDaniel, ``Ensemble adversarial training: Attacks and defenses,'' in
  \emph{6th International Conference on Learning Representations, {ICLR} 2018,
  Vancouver, BC, Canada, April 30 - May 3, 2018, Conference Track
  Proceedings}.\hskip 1em plus 0.5em minus 0.4em\relax OpenReview.net, 2018.

\bibitem{carlini2017towards}
N.~Carlini and D.~A. Wagner, ``Towards evaluating the robustness of neural
  networks,'' in \emph{2017 {IEEE} Symposium on Security and Privacy, {SP}
  2017, San Jose, CA, USA, May 22-26, 2017}.\hskip 1em plus 0.5em minus
  0.4em\relax {IEEE} Computer Society, 2017, pp. 39--57.

\bibitem{dong2019evading}
Y.~Dong, T.~Pang, H.~Su, and J.~Zhu, ``Evading defenses to transferable
  adversarial examples by translation-invariant attacks,'' in \emph{{IEEE}
  Conference on Computer Vision and Pattern Recognition, {CVPR} 2019, Long
  Beach, CA, USA, June 16-20, 2019}.\hskip 1em plus 0.5em minus 0.4em\relax
  Computer Vision Foundation / {IEEE}, 2019, pp. 4312--4321.

\bibitem{lin2019nesterov}
J.~Lin, C.~Song, K.~He, L.~Wang, and J.~E. Hopcroft, ``Nesterov accelerated
  gradient and scale invariance for adversarial attacks,'' 2020.

\bibitem{xie2019improving}
C.~Xie, Z.~Zhang, Y.~Zhou, S.~Bai, J.~Wang, Z.~Ren, and A.~L. Yuille,
  ``Improving transferability of adversarial examples with input diversity,''
  in \emph{{IEEE} Conference on Computer Vision and Pattern Recognition, {CVPR}
  2019, Long Beach, CA, USA, June 16-20, 2019}.\hskip 1em plus 0.5em minus
  0.4em\relax Computer Vision Foundation / {IEEE}, 2019, pp. 2730--2739.

\bibitem{brendel2017decision}
W.~Brendel, J.~Rauber, and M.~Bethge, ``Decision-based adversarial attacks:
  Reliable attacks against black-box machine learning models,'' in \emph{6th
  International Conference on Learning Representations, {ICLR} 2018, Vancouver,
  BC, Canada, April 30 - May 3, 2018, Conference Track Proceedings}.\hskip 1em
  plus 0.5em minus 0.4em\relax OpenReview.net, 2018.

\bibitem{ilyas2018black}
A.~Ilyas, L.~Engstrom, A.~Athalye, and J.~Lin, ``Black-box adversarial attacks
  with limited queries and information,'' in \emph{Proceedings of the 35th
  International Conference on Machine Learning, {ICML} 2018,
  Stockholmsm{\"{a}}ssan, Stockholm, Sweden, July 10-15, 2018}, ser.
  Proceedings of Machine Learning Research, J.~G. Dy and A.~Krause, Eds.,
  vol.~80.\hskip 1em plus 0.5em minus 0.4em\relax {PMLR}, 2018, pp. 2142--2151.

\bibitem{chen2020boosting}
W.~Chen, Z.~Zhang, X.~Hu, and B.~Wu, ``Boosting decision-based black-box
  adversarial attacks with random sign flip,'' in \emph{Computer Vision -
  {ECCV} 2020 - 16th European Conference, Glasgow, UK, August 23-28, 2020,
  Proceedings, Part {XV}}, ser. Lecture Notes in Computer Science, A.~Vedaldi,
  H.~Bischof, T.~Brox, and J.~Frahm, Eds., vol. 12360.\hskip 1em plus 0.5em
  minus 0.4em\relax Springer, 2020, pp. 276--293.

\bibitem{croce2020reliable}
F.~Croce and M.~Hein, ``Reliable evaluation of adversarial robustness with an
  ensemble of diverse parameter-free attacks,'' in \emph{Proceedings of the
  37th International Conference on Machine Learning, {ICML} 2020, 13-18 July
  2020, Virtual Event}, ser. Proceedings of Machine Learning Research, vol.
  119.\hskip 1em plus 0.5em minus 0.4em\relax {PMLR}, 2020, pp. 2206--2216.

\bibitem{croce2020minimally}
------, ``Minimally distorted adversarial examples with a fast adaptive
  boundary attack,'' in \emph{Proceedings of the 37th International Conference
  on Machine Learning, {ICML} 2020, 13-18 July 2020, Virtual Event}, ser.
  Proceedings of Machine Learning Research, vol. 119.\hskip 1em plus 0.5em
  minus 0.4em\relax {PMLR}, 2020, pp. 2196--2205.

\bibitem{andriushchenko2020square}
M.~Andriushchenko, F.~Croce, N.~Flammarion, and M.~Hein, ``Square attack: {A}
  query-efficient black-box adversarial attack via random search,'' in
  \emph{Computer Vision - {ECCV} 2020 - 16th European Conference, Glasgow, UK,
  August 23-28, 2020, Proceedings, Part {XXIII}}, ser. Lecture Notes in
  Computer Science, A.~Vedaldi, H.~Bischof, T.~Brox, and J.~Frahm, Eds., vol.
  12368.\hskip 1em plus 0.5em minus 0.4em\relax Springer, 2020, pp. 484--501.

\bibitem{zhang2019theoretically}
H.~Zhang, Y.~Yu, J.~Jiao, E.~P. Xing, L.~E. Ghaoui, and M.~I. Jordan,
  ``Theoretically principled trade-off between robustness and accuracy,'' in
  \emph{Proceedings of the 36th International Conference on Machine Learning,
  {ICML} 2019, 9-15 June 2019, Long Beach, California, {USA}}, ser. Proceedings
  of Machine Learning Research, K.~Chaudhuri and R.~Salakhutdinov, Eds.,
  vol.~97.\hskip 1em plus 0.5em minus 0.4em\relax {PMLR}, 2019, pp. 7472--7482.

\bibitem{wang2019improving}
Y.~Wang, D.~Zou, J.~Yi, J.~Bailey, X.~Ma, and Q.~Gu, ``Improving adversarial
  robustness requires revisiting misclassified examples,'' in \emph{8th
  International Conference on Learning Representations, {ICLR} 2020, Addis
  Ababa, Ethiopia, April 26-30, 2020}.\hskip 1em plus 0.5em minus 0.4em\relax
  OpenReview.net, 2020.

\bibitem{roth2019adversarial}
K.~Roth, Y.~Kilcher, and T.~Hofmann, ``Adversarial training is a form of
  data-dependent operator norm regularization,'' in \emph{Advances in Neural
  Information Processing Systems 33: Annual Conference on Neural Information
  Processing Systems 2020, NeurIPS 2020, December 6-12, 2020, virtual},
  H.~Larochelle, M.~Ranzato, R.~Hadsell, M.~Balcan, and H.~Lin, Eds., 2020.

\bibitem{DBLP:conf/icml/Yu0SY0GL22}
C.~Yu, B.~Han, L.~Shen, J.~Yu, C.~Gong, M.~Gong, and T.~Liu, ``Understanding
  robust overfitting of adversarial training and beyond,'' in
  \emph{International Conference on Machine Learning, {ICML} 2022, 17-23 July
  2022, Baltimore, Maryland, {USA}}, ser. Proceedings of Machine Learning
  Research, K.~Chaudhuri, S.~Jegelka, L.~Song, C.~Szepesv{\'{a}}ri, G.~Niu, and
  S.~Sabato, Eds., vol. 162.\hskip 1em plus 0.5em minus 0.4em\relax {PMLR},
  2022, pp. 25\,595--25\,610.

\bibitem{lee2020adversarial}
S.~Lee, H.~Lee, and S.~Yoon, ``Adversarial vertex mixup: Toward better
  adversarially robust generalization,'' in \emph{2020 {IEEE/CVF} Conference on
  Computer Vision and Pattern Recognition, {CVPR} 2020, Seattle, WA, USA, June
  13-19, 2020}.\hskip 1em plus 0.5em minus 0.4em\relax Computer Vision
  Foundation / {IEEE}, 2020, pp. 269--278.

\bibitem{wang2021convergence}
Y.~Wang, X.~Ma, J.~Bailey, J.~Yi, B.~Zhou, and Q.~Gu, ``On the convergence and
  robustness of adversarial training,'' in \emph{Proceedings of the 36th
  International Conference on Machine Learning, {ICML} 2019, 9-15 June 2019,
  Long Beach, California, {USA}}, ser. Proceedings of Machine Learning
  Research, K.~Chaudhuri and R.~Salakhutdinov, Eds., vol.~97.\hskip 1em plus
  0.5em minus 0.4em\relax {PMLR}, 2019, pp. 6586--6595.

\bibitem{shafahi2019adversarial}
A.~Shafahi, M.~Najibi, A.~Ghiasi, Z.~Xu, J.~P. Dickerson, C.~Studer, L.~S.
  Davis, G.~Taylor, and T.~Goldstein, ``Adversarial training for free!'' in
  \emph{Advances in Neural Information Processing Systems 32: Annual Conference
  on Neural Information Processing Systems 2019, NeurIPS 2019, December 8-14,
  2019, Vancouver, BC, Canada}, H.~M. Wallach, H.~Larochelle, A.~Beygelzimer,
  F.~d'Alch{\'{e}}{-}Buc, E.~B. Fox, and R.~Garnett, Eds., 2019, pp.
  3353--3364.

\bibitem{dhillon2018stochastic}
G.~S. Dhillon, K.~Azizzadenesheli, Z.~C. Lipton, J.~Bernstein, J.~Kossaifi,
  A.~Khanna, and A.~Anandkumar, ``Stochastic activation pruning for robust
  adversarial defense,'' in \emph{6th International Conference on Learning
  Representations, {ICLR} 2018, Vancouver, BC, Canada, April 30 - May 3, 2018,
  Conference Track Proceedings}.\hskip 1em plus 0.5em minus 0.4em\relax
  OpenReview.net, 2018.

\bibitem{li2020towards}
B.~Li, S.~Wang, Y.~Jia, Y.~Lu, Z.~Zhong, L.~Carin, and S.~Jana, ``Towards
  practical lottery ticket hypothesis for adversarial training,'' \emph{arXiv
  preprint arXiv:2003.05733}, 2020.

\bibitem{frankle2018lottery}
J.~Frankle and M.~Carbin, ``The lottery ticket hypothesis: Finding sparse,
  trainable neural networks,'' in \emph{7th International Conference on
  Learning Representations, {ICLR} 2019, New Orleans, LA, USA, May 6-9,
  2019}.\hskip 1em plus 0.5em minus 0.4em\relax OpenReview.net, 2019.

\bibitem{he2020robustness}
F.~He, S.~Fu, B.~Wang, and D.~Tao, ``Robustness, privacy, and generalization of
  adversarial training,'' \emph{arXiv preprint arXiv:2012.13573}, 2020.

\bibitem{huang2016deep}
G.~Huang, Y.~Sun, Z.~Liu, D.~Sedra, and K.~Q. Weinberger, ``Deep networks with
  stochastic depth,'' in \emph{Computer Vision - {ECCV} 2016 - 14th European
  Conference, Amsterdam, The Netherlands, October 11-14, 2016, Proceedings,
  Part {IV}}, ser. Lecture Notes in Computer Science, B.~Leibe, J.~Matas,
  N.~Sebe, and M.~Welling, Eds., vol. 9908.\hskip 1em plus 0.5em minus
  0.4em\relax Springer, 2016, pp. 646--661.

\bibitem{yang2020rethinking}
Z.~Yang, Y.~Yu, C.~You, J.~Steinhardt, and Y.~Ma, ``Rethinking bias-variance
  trade-off for generalization of neural networks,'' in \emph{Proceedings of
  the 37th International Conference on Machine Learning, {ICML} 2020, 13-18
  July 2020, Virtual Event}, ser. Proceedings of Machine Learning Research,
  vol. 119.\hskip 1em plus 0.5em minus 0.4em\relax {PMLR}, 2020, pp.
  10\,767--10\,777.

\bibitem{hayou2021regularization}
S.~Hayou and F.~Ayed, ``Regularization in resnet with stochastic depth,'' in
  \emph{Advances in Neural Information Processing Systems 34: Annual Conference
  on Neural Information Processing Systems 2021, NeurIPS 2021, December 6-14,
  2021, virtual}, M.~Ranzato, A.~Beygelzimer, Y.~N. Dauphin, P.~Liang, and
  J.~W. Vaughan, Eds., 2021, pp. 15\,464--15\,474.

\bibitem{krizhevsky2009learning}
A.~Krizhevsky, G.~Hinton \emph{et~al.}, ``Learning multiple layers of features
  from tiny images.''\hskip 1em plus 0.5em minus 0.4em\relax Citeseer, 2009.

\bibitem{deng2009imagenet}
J.~Deng, W.~Dong, R.~Socher, L.~Li, K.~Li, and L.~Fei{-}Fei, ``Imagenet: {A}
  large-scale hierarchical image database,'' in \emph{2009 {IEEE} Computer
  Society Conference on Computer Vision and Pattern Recognition {(CVPR} 2009),
  20-25 June 2009, Miami, Florida, {USA}}.\hskip 1em plus 0.5em minus
  0.4em\relax {IEEE} Computer Society, 2009, pp. 248--255.

\bibitem{rice2020overfitting}
L.~Rice, E.~Wong, and J.~Z. Kolter, ``Overfitting in adversarially robust deep
  learning,'' in \emph{Proceedings of the 37th International Conference on
  Machine Learning, {ICML} 2020, 13-18 July 2020, Virtual Event}, ser.
  Proceedings of Machine Learning Research, vol. 119.\hskip 1em plus 0.5em
  minus 0.4em\relax {PMLR}, 2020, pp. 8093--8104.

\bibitem{he2016deep}
K.~He, X.~Zhang, S.~Ren, and J.~Sun, ``Deep residual learning for image
  recognition,'' in \emph{2016 {IEEE} Conference on Computer Vision and Pattern
  Recognition, {CVPR} 2016, Las Vegas, NV, USA, June 27-30, 2016}.\hskip 1em
  plus 0.5em minus 0.4em\relax {IEEE} Computer Society, 2016, pp. 770--778.

\bibitem{he2016identity}
------, ``Identity mappings in deep residual networks,'' in \emph{Computer
  Vision - {ECCV} 2016 - 14th European Conference, Amsterdam, The Netherlands,
  October 11-14, 2016, Proceedings, Part {IV}}, ser. Lecture Notes in Computer
  Science, B.~Leibe, J.~Matas, N.~Sebe, and M.~Welling, Eds., vol. 9908.\hskip
  1em plus 0.5em minus 0.4em\relax Springer, 2016, pp. 630--645.

\bibitem{pang2020bag}
T.~Pang, X.~Yang, Y.~Dong, H.~Su, and J.~Zhu, ``Bag of tricks for adversarial
  training,'' in \emph{9th International Conference on Learning
  Representations, {ICLR} 2021, Virtual Event, Austria, May 3-7, 2021}.\hskip
  1em plus 0.5em minus 0.4em\relax OpenReview.net, 2021.

\bibitem{DBLP:conf/nips/WuX020}
D.~Wu, S.~Xia, and Y.~Wang, ``Adversarial weight perturbation helps robust
  generalization,'' in \emph{Advances in Neural Information Processing Systems
  33: Annual Conference on Neural Information Processing Systems 2020, NeurIPS
  2020, December 6-12, 2020, virtual}, H.~Larochelle, M.~Ranzato, R.~Hadsell,
  M.~Balcan, and H.~Lin, Eds., 2020.

\end{thebibliography}










\newpage

 




\vfill

\end{document}